\PassOptionsToPackage{hyphens}{url}
\PassOptionsToPackage{most}{tcolorbox}
\PassOptionsToPackage{table}{xcolor}
\RequirePackage{fix-cm}
\documentclass[11pt]{article}
\let\standardaddcontentsline\addcontentsline
\usepackage[preprint]{arxiv}
\let\addcontentsline\standardaddcontentsline
\usepackage{times}
\usepackage[utf8]{inputenc}
\usepackage[T1]{fontenc}
\usepackage{microtype}

\usepackage{amsmath,amsfonts,bm}

\def\eqref#1{equation~\ref{#1}}

\def\1{\bm{1}}

\DeclareMathAlphabet{\mathsfit}{\encodingdefault}{\sfdefault}{m}{sl}
\SetMathAlphabet{\mathsfit}{bold}{\encodingdefault}{\sfdefault}{bx}{n}

\usepackage{url}
\usepackage{booktabs}
\usepackage{tabularx}
\usepackage{graphicx}
\usepackage{xcolor}
\usepackage{multirow}
\usepackage{etoolbox}
\usepackage{float}
\usepackage{xspace}
\usepackage{subcaption}
\usepackage{wrapfig}
\usepackage{fvextra}
\usepackage{placeins}
\usepackage{needspace}
\usepackage[most]{tcolorbox}
\BeforeBeginEnvironment{abstract}{\begingroup\tcbset{breakable}}
\AfterEndEnvironment{abstract}{\endgroup}
\hypersetup{
  hypertexnames=false,
  pdftitle={YANAMI: A Large-Scale Multimodal Dataset and Benchmark for Multiturn Stateful Interaction},
  pdfauthor={Shengyu Wang, Tengyue Nong, Junming Liu, Haosong Peng, Siyu Li, Ruiqi Lei, Fushuo Huo, Yuhao Pan, Yunfeng Fan, Ding Wang, Nan Cheng, Haozhao Wang, Wenchao Xu}
}

\title{YANAMI: A Large-Scale Multimodal Dataset and Benchmark for Multiturn Stateful Interaction}

\author{\parbox{\textwidth}{\centering\hypersetup{urlcolor=black}\normalfont\fontsize{10}{12}\selectfont
  \mbox{\href{mailto:swanghh@connect.ust.hk}{Shengyu Wang}\textsuperscript{1*\textdagger}},
  \mbox{\href{mailto:rowletqwq@foxmail.com}{Tengyue Nong}\textsuperscript{2*}},
  \mbox{\href{mailto:pppplongke@gmail.com}{Junming Liu}\textsuperscript{3*}},
  \mbox{\href{mailto:livion_i@icloud.com}{Haosong Peng}\textsuperscript{1*}},
  \mbox{\href{mailto:lys1774246905@gmail.com}{Siyu Li}\textsuperscript{4*}},\\
  \mbox{\href{mailto:ruiqi.lei25@outlook.com}{Ruiqi Lei}\textsuperscript{5}},
  \mbox{\href{mailto:fushuohuo@seu.edu.cn}{Fushuo Huo}\textsuperscript{6}},
  \mbox{\href{mailto:ypanca@connect.ust.hk}{Yuhao Pan}\textsuperscript{1}},
  \mbox{\href{mailto:fan_yun_feng@163.com}{Yunfeng Fan}\textsuperscript{7}},\\
  \mbox{\href{mailto:wangding.mellissa@gmail.com}{Ding Wang}\textsuperscript{8}},
  \mbox{\href{mailto:nancheng@xidian.edu.cn}{Nan Cheng}\textsuperscript{9}},
  \mbox{\href{mailto:hz_wang@hust.edu.cn}{Haozhao Wang}\textsuperscript{10}},
  \mbox{\href{mailto:wenchaoxu@ust.hk}{Wenchao Xu}\textsuperscript{1\ensuremath{\ddagger}}}\endgraf
  \medskip
  {\footnotesize
    \mbox{\textsuperscript{1}\,The Hong Kong University of Science and Technology}\quad
    \mbox{\textsuperscript{2}\,Independent Researcher}\quad
    \mbox{\textsuperscript{3}\,The University of Melbourne}\quad
    \mbox{\textsuperscript{4}\,The Chinese University of Hong Kong, Shenzhen}\quad
    \mbox{\textsuperscript{5}\,Imperial College London}\quad
    \mbox{\textsuperscript{6}\,Southeast University}\quad
    \mbox{\textsuperscript{7}\,The Hong Kong Polytechnic University}\quad
    \mbox{\textsuperscript{8}\,Shanghai Artificial Intelligence Laboratory}\quad
    \mbox{\textsuperscript{9}\,Xidian University}\quad
    \mbox{\textsuperscript{10}\,Huazhong University of Science and Technology}\endgraf
    \smallskip
    \textbf{\textcolor[HTML]{D94B62}{Project}\textcolor[HTML]{DE8535}{\ page:}}\ %
    \href{https://rowletqwq.github.io/YANAMI}{\texttt{%
      \textcolor[HTML]{D94B62}{https:}%
      \textcolor[HTML]{DE8535}{//row}%
      \textcolor[HTML]{B49A22}{letqw}%
      \textcolor[HTML]{38965B}{q.git}%
      \textcolor[HTML]{269EA8}{hub.}%
      \textcolor[HTML]{4879CA}{io/YA}%
      \textcolor[HTML]{9758BB}{NAMI}}}\endgraf
  }%
}}

\newcommand{\dialoguename}{YANAMI\xspace}
\newcommand{\dialoguedataset}{YANAMI-400K\xspace}
\newcommand{\dialoguebench}{YANAMI-Bench\xspace}
\newlength{\dialoguepanelheight}
\DeclareRobustCommand{\tableicon}[1]{%
  \raisebox{-0.19\height}{\includegraphics[height=1.45em]{figures/icon/#1.png}}%
}

\begin{document}

\setlength{\headheight}{24pt}
\fancyhf{}
\cfoot{\thepage}
\renewcommand{\headrulewidth}{0.4pt}
\fancypagestyle{firstpage}{
  \fancyhf{}
  \lhead{%
    \includegraphics[height=18pt]{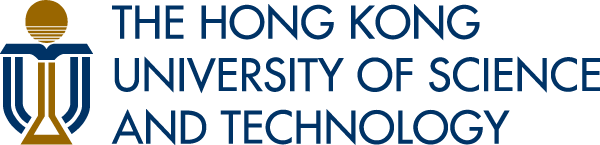}%
    \hspace{0.7em}%
    \includegraphics[height=18pt]{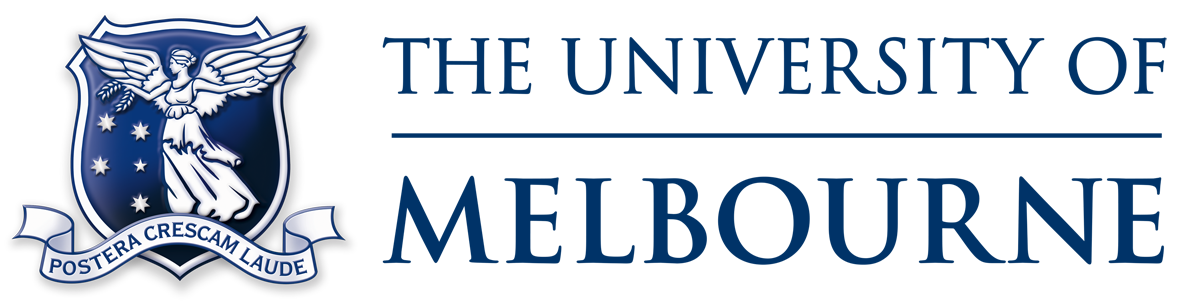}%
    \hspace{0.7em}%
    \includegraphics[height=18pt]{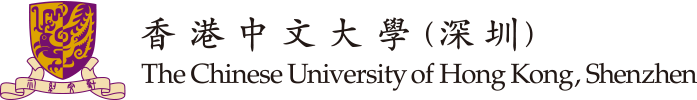}%
    \hspace{0.7em}%
    \includegraphics[height=18pt]{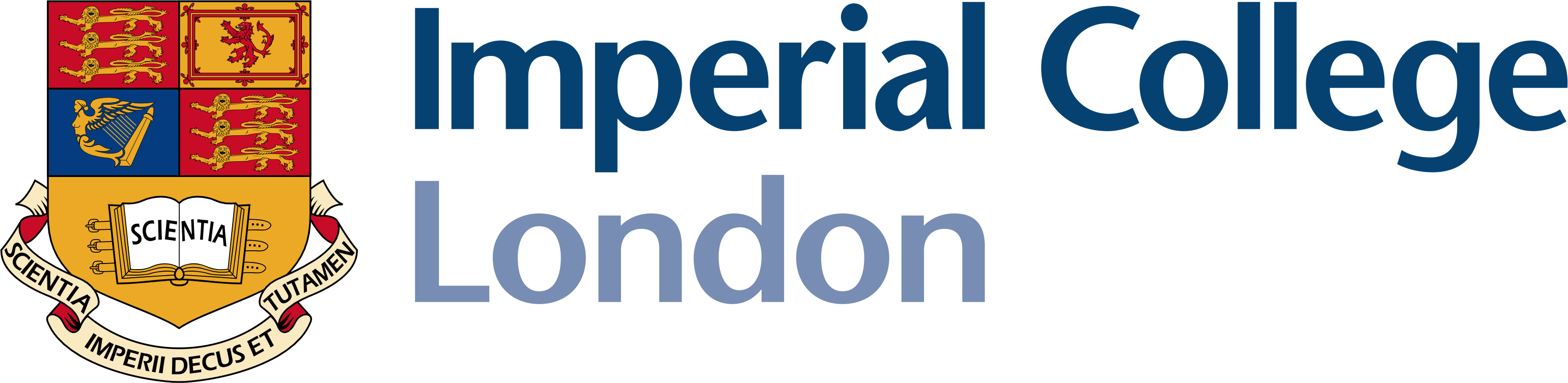}%
  }
  \rhead{\small\textcolor[HTML]{4D4D4D}{\today\quad Preprint}}
  \cfoot{\thepage}
  \renewcommand{\headrulewidth}{0.4pt}
}

\maketitle
\begingroup
  \renewcommand{\thefootnote}{\fnsymbol{footnote}}
  \footnotetext[1]{Equal contribution: Shengyu Wang, Tengyue Nong, Junming Liu, Haosong Peng, and Siyu Li.}
  \footnotetext[2]{Project Leader: Shengyu Wang (\href{mailto:swanghh@connect.ust.hk}{\texttt{swanghh@connect.ust.hk}}).}
  \footnotetext[3]{Corresponding author: Wenchao Xu (\href{mailto:wenchaoxu@ust.hk}{\texttt{wenchaoxu@ust.hk}}).}
\endgroup
\vspace{-10mm}
\begin{figure}[H]
  \centering
  \includegraphics[width=\linewidth]{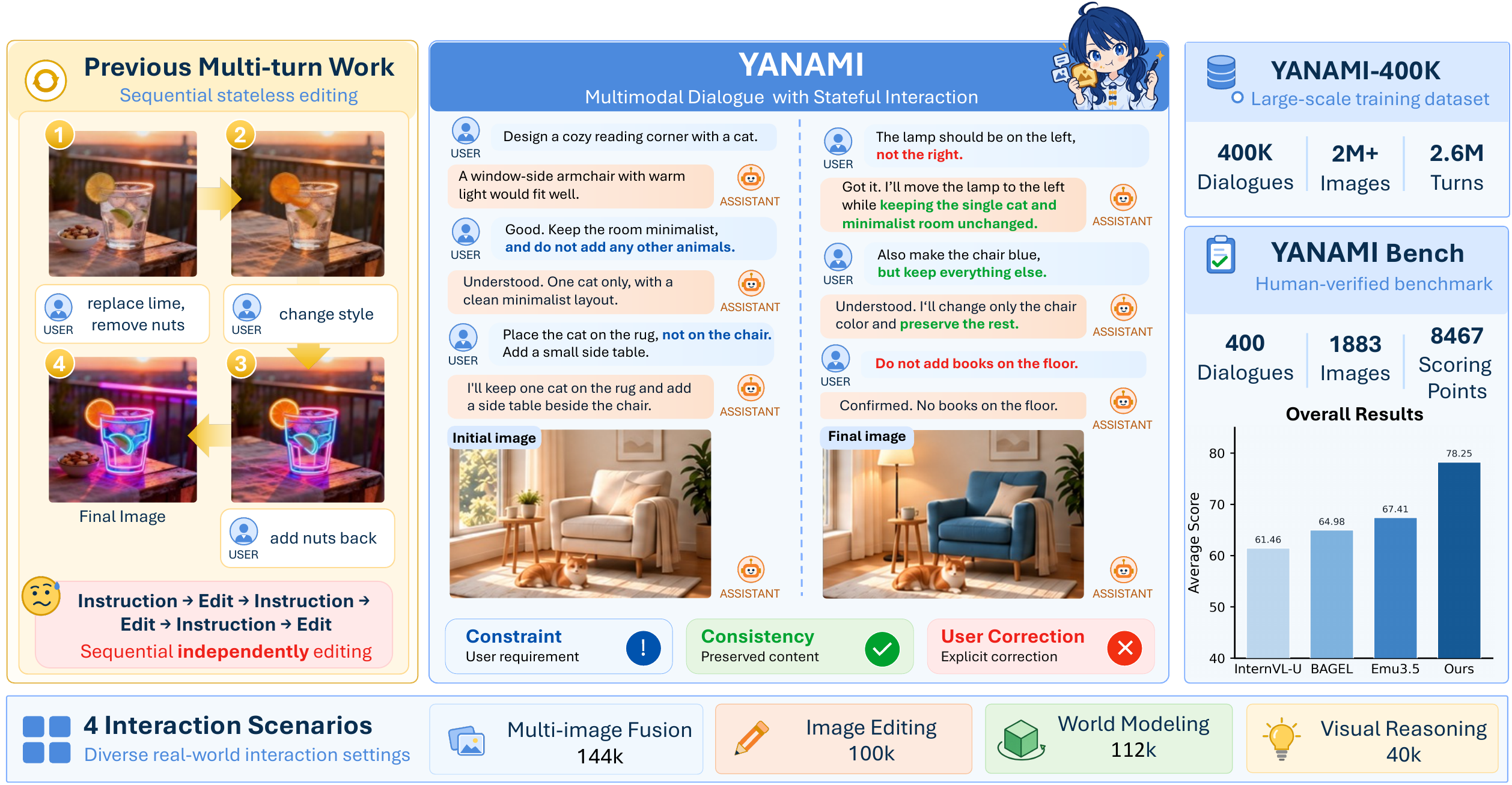}
  \caption{\textbf{\dialoguename:} the first large-scale multimodal dialogue dataset and benchmark for stateful interaction.}
  \label{fig:teaser}
\end{figure}

\begin{abstract}
Unified Multimodal Models (UMMs) integrate visual understanding, generation, and editing within a single unified framework.
However, existing unified multimodal datasets and benchmarks remain dominated by single-turn tasks.
Even in recent multi-turn datasets, many turns can be solved independently without tracking information across turns.
Consequently, they provide limited supervision and evaluation for multimodal \emph{stateful interaction}, which requires models to understand, maintain, update, and faithfully realize evolving dialogue states across turns.
To address this gap, we introduce \textbf{\dialoguename} (\textbf{\underline{Y}}our \textbf{\underline{A}}ssistant \textbf{\underline{N}}eeds to \textbf{\underline{A}}ssimilate \textbf{\underline{M}}ultiturn \textbf{\underline{I}}nteraction), the first large-scale multimodal dialogue dataset and benchmark for stateful interaction, comprising \textbf{\dialoguedataset} and \textbf{\dialoguebench}.
\dialoguedataset contains 400K multi-turn multimodal dialogues, with more than 2M images and 2.6M dialogue turns.
The dialogues span four representative scenarios, Multi-Image Fusion, Image Editing, World Modeling, and Visual Reasoning, and require models to continuously track, preserve, and revise evolving multimodal dialogue states across turns.
\dialoguebench comprises 400 multi-turn dialogues for systematically evaluating multimodal stateful interaction.
It jointly evaluates \textbf{Dialogue State Understanding} and \textbf{Visual Realization}, characterizing the gap between models' ability to understand evolving dialogue states and their ability to faithfully realize them in visual outputs.
Experiments show that fine-tuning BAGEL on \dialoguedataset{} improves image editing, generation, and multimodal stateful interaction by 10.3\%, 2.0\%, and 13.23\%, respectively, while preserving its general visual understanding capabilities.
Evaluations on \dialoguebench{} further reveal that existing UMMs still struggle to maintain, update, and visually realize evolving multimodal dialogue states.
Together, \dialoguename{} provides a unified foundation for training and evaluating UMMs toward robust multimodal stateful interaction.

\end{abstract}

\section{introduction}
\vspace{-6pt}
Unified Multimodal Models (UMMs)~\citep{deng2025emerging,wu2025omnigen2,xie2026show} represent an emerging shift from task-specific systems toward unified visual modeling.
Unlike conventional models designed for individual tasks~\citep{liu2023visual,rombach2022high}, UMMs integrate image understanding, generation, and editing within a single framework. This unified modeling enables UMMs to interleave processing across textual and visual modalities, highlighting their potential for complex multi-turn multimodal interaction.

However, most publicly available datasets focus on single-turn tasks \citep{chen2025blip3,wang2025gpt}.
Only a few support multi-turn interaction ~\citep{chow2026weave,feng2026high}, and even these often contain turns that can be solved \emph{independently} without tracking information across turns, falling short of complex multi-turn multimodal interaction.
In contrast, real-world user-assistant image creation and editing are typically not completed in a single step, but instead involve continuous multi-turn \emph{stateful interaction} for gradual refinement.

Here, \emph{stateful interaction} refers to a multi-turn dialogue in which the user and assistant discuss alternatives, make decisions, establish cross-turn constraints and dependencies, plan subsequent steps, and revise previous decisions or requirements in response to user corrections.
Consequently, datasets without such interactions provide limited supervision for learning to understand, maintain, update, and faithfully realize evolving multimodal \emph{dialogue states} across turns.
It further limits the exploration of UMMs for complex multi-turn multimodal interaction.
For example, as illustrated in Fig.~\ref{fig:teaser}, existing multi-turn datasets typically alternate between user instructions and assistant edits, with each turn treated independently and lacking cross-turn contextual dependencies, whereas our dataset captures stateful user–assistant interactions in which dialogue states are established, maintained, and updated through ongoing context-dependent exchanges.

To address these limitations, we introduce \textbf{\dialoguename}, the first large-scale multimodal dialogue dataset and benchmark for \emph{stateful interaction}, comprising \textbf{\dialoguedataset} and \textbf{\dialoguebench} to capture evolving multimodal dialogue states and evaluate models' performance across understanding, generation, and editing.
\dialoguedataset contains \textbf{400K} multi-turn multimodal dialogues, comprising more than \textbf{2M} images and \textbf{2.6M} dialogue turns in total.
To construct genuinely stateful interactions, we explicitly incorporate six key capabilities into each dialogue: \textit{1) Reference Resolution}; \textit{2) Context Retrieval}; \textit{3) Consistency Maintenance}; \textit{4) Spatial and Physical Consistency}; \textit{5) Constraint Keeping}; and \textit{6) Correction-Based State Updating}. 
Together, these capabilities make stateful interaction explicit in the data, with each dialogue requiring multimodal states to be continuously tracked, preserved, and revised across turns.
To instantiate such stateful interaction in diverse settings, we construct dialogues under four representative scenarios: \textit{1) Multi-Image Fusion}; \textit{2) Image Editing}; \textit{3) World Modeling}; and \textit{4) Visual Reasoning}.
These scenarios jointly exercise diverse capabilities, spanning visual-state composition, modification, transition, and reasoning.

\dialoguebench comprises 400 multi-turn dialogues designed to systematically evaluate models' ability to track, maintain, and update multimodal dialogue states throughout interactions.
Specifically, \dialoguebench covers the same four scenarios, including complex understanding, generation, and editing tasks, with each dialogue jointly involving all six key capabilities and requiring their coordinated use throughout the interaction.
\dialoguebench evaluates models along two complementary dimensions: \textbf{Dialogue State Understanding (DSU)} and \textbf{Visual Realization (VR)}.
\textbf{DSU} evaluates whether a model can infer the currently valid multimodal state from the full interaction history.
We assess it through capability-oriented dialogue-based visual question answering along four dimensions: \textit{1) Reference Resolution}; \textit{2) Context Retrieval}; \textit{3) Constraint Keeping}; and \textit{4) Correction-Based State Updating}.
VR measures whether the generated images faithfully realize the current dialogue state and is assessed by \textit{Visual Quality} and \textit{State Consistency}.
Together, these two dimensions provide a unified evaluation of stateful multimodal interaction.

Our experiments demonstrate that training on \dialoguedataset substantially improves model performance on image generation, image editing, and multi-turn stateful interaction tasks while maintaining general visual understanding capabilities.
Meanwhile, evaluations on \dialoguebench reveal several key limitations of current models in long-horizon multimodal interactions.
We hope that \dialoguename provides a unified foundation for training and evaluating UMMs toward robust multimodal stateful interaction.
In summary, our main contributions are as follows:
\begin{itemize}

\item We introduce \dialoguedataset, the first large-scale multimodal dialogue dataset for stateful interaction across understanding, generation, and editing.
It comprises \textbf{400K} multi-turn dialogues with more than \textbf{2M} images and \textbf{2.6M} dialogue turns, providing large-scale supervision for six key capabilities required for stateful interaction.

\item We construct \dialoguebench, a benchmark of 400 multi-turn dialogues.
By jointly evaluating DSU and VR, \dialoguebench characterizes the gap between models' ability to understand dialogue states and their ability to faithfully realize them in visual outputs.

\item Fine-tuning BAGEL~\citep{deng2025emerging} on \dialoguedataset yields consistent gains in general image generation and editing, as well as DSU and VR, while preserving general visual understanding capabilities.
Results on \dialoguebench reveal that existing UMMs still face substantial challenges in maintaining, updating, and visually realizing evolving multimodal dialogue states.

\end{itemize}

\section{related work}
\label{related work}
\textbf{Unified Multimodal Models} aim to integrate visual understanding and generation within a shared architecture, providing a unified interface for processing and producing multimodal content.
Existing UMMs explore diverse architectural paradigms, ranging from unified autoregressive modeling\citep{team2024chameleon,wu2025janus} over multimodal representations to hybrid frameworks that combine autoregressive language modeling with diffusion- or flow-based visual generation~\citep{deng2025emerging,wu2025omnigen2}.
Existing UMMs have demonstrated increasing capabilities across single-turn multimodal understanding, generation, and editing tasks \citep{chen2025janus,cui2025emu3}.
However, how well these capabilities extend to multi-turn multimodal dialogues with evolving contextual dependencies remains largely underexplored.

\textbf{Multimodal Data and Benchmarks for Understanding, Generation, and Editing.}
Existing multimodal datasets primarily provide single-turn supervision for individual capabilities such as understanding~\citep{liu2023visual}, generation~\citep{brooks2023instructpix2pix}, and editing~\citep{yu2025anyedit}, with some extending to multi-image settings~\citep{chen2025comm,li2025zebra}.
Recent work has further explored multi-turn interaction through sequential editing and interleaved dialogues~\citep{chow2026weave,qian2026pico,dai2026chatumm}.
Meanwhile, existing benchmarks largely evaluate individual capabilities in isolation, including understanding~\citep{fu2026mme,liu2024mmbench,li2023evaluating}, generation~\citep{ghosh2023geneval}, and editing~\citep{liu2025step1x}.
More recent efforts have begun to evaluate unified or interleaved capabilities~\citep{xia2025mmie,xie2026mme} and multi-turn settings spanning understanding, editing, generation, and unified multimodal interaction~\citep{liu2024convbench,chen2025edival,huang2025dialoggen,chow2026weave}.
However, these datasets and benchmarks can often be approached turn by turn without explicitly tracking evolving information across the dialogue, limiting their ability to support and evaluate truly multimodal stateful interaction.
In contrast, \dialoguename{} jointly provides large-scale supervision and systematic evaluation for tracking, updating, and faithfully realizing evolving multimodal dialogue states across turns.
The comparison between \dialoguename{} and multimodal datasets is provided in Appendix~\ref{app:dataset_comparison}.

\section{\textbf{\dialoguename{}}}
\label{dialogue}
In this section, we present \dialoguename, our dataset and benchmark for multimodal stateful interaction.
\dialoguedataset provides large-scale training data, while \dialoguebench evaluates models across understanding, generation, and editing.
We first describe the data construction process in Sec. \ref{data_collection}, followed by the evaluation setting and metrics in Sec. \ref{eval_setting} and the dataset statistics in Sec. \ref{data_statics}.

\begin{figure}[thbp]
    \centering
    \includegraphics[width=\linewidth,height=0.87\textheight,keepaspectratio]{
        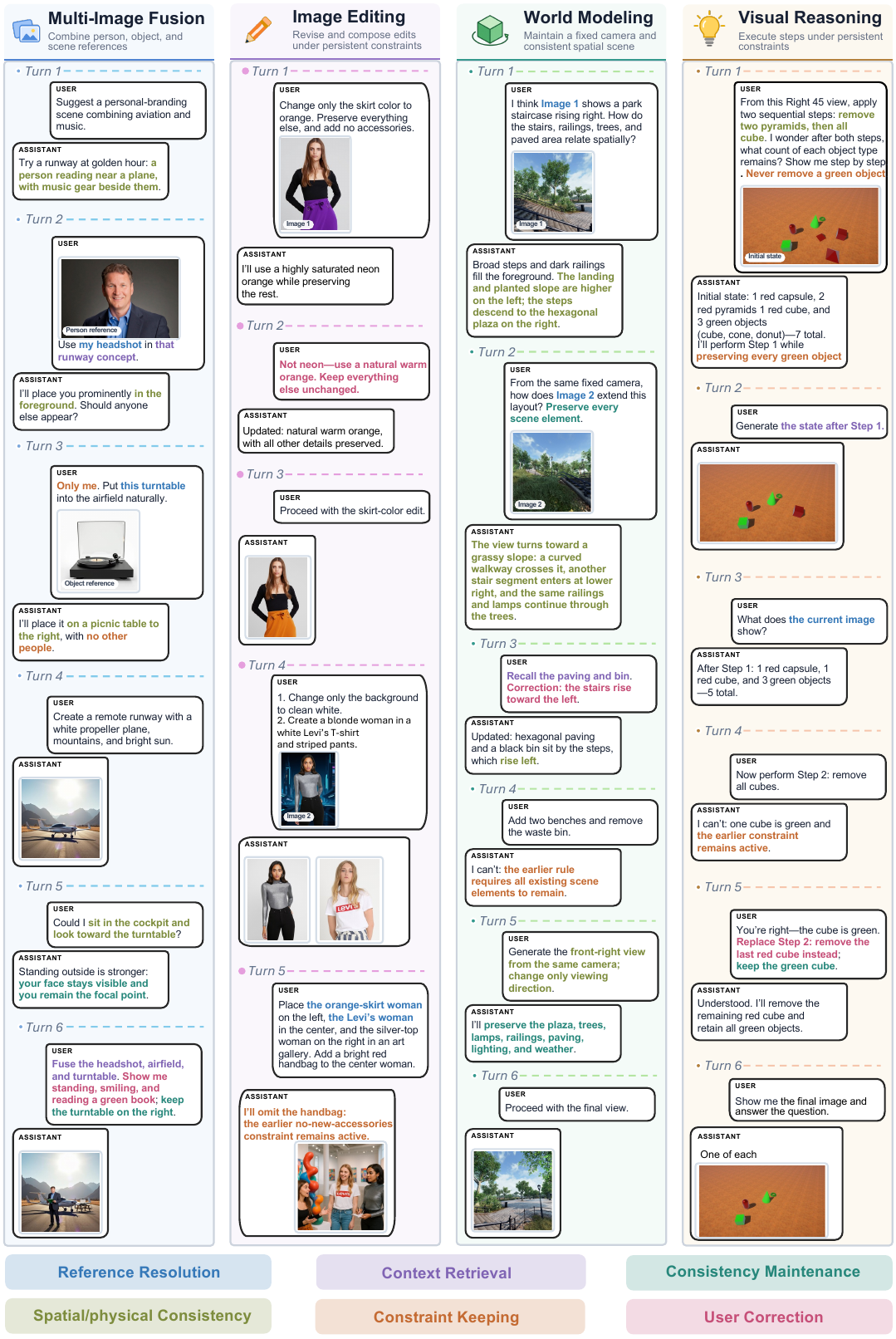
    }
   \caption{\textbf{Representative examples from \dialoguedataset{}.}
The four columns present evolving multimodal dialogues in four interaction scenarios: (a) multi-image fusion, (b) image editing, (c) world modeling, and (d) visual reasoning.
Across turns, subsequent responses retrieve, maintain, and update the accumulated dialogue state.
Colored text highlights six key capabilities.}
\label{fig:dataset_examples}
    \label{fig:dataset-examples-four-columns}
\end{figure}

\subsection{Data Pipeline}
\label{data_collection}

\begin{figure}[thbp]
    \centering
    \includegraphics[width=1\linewidth]{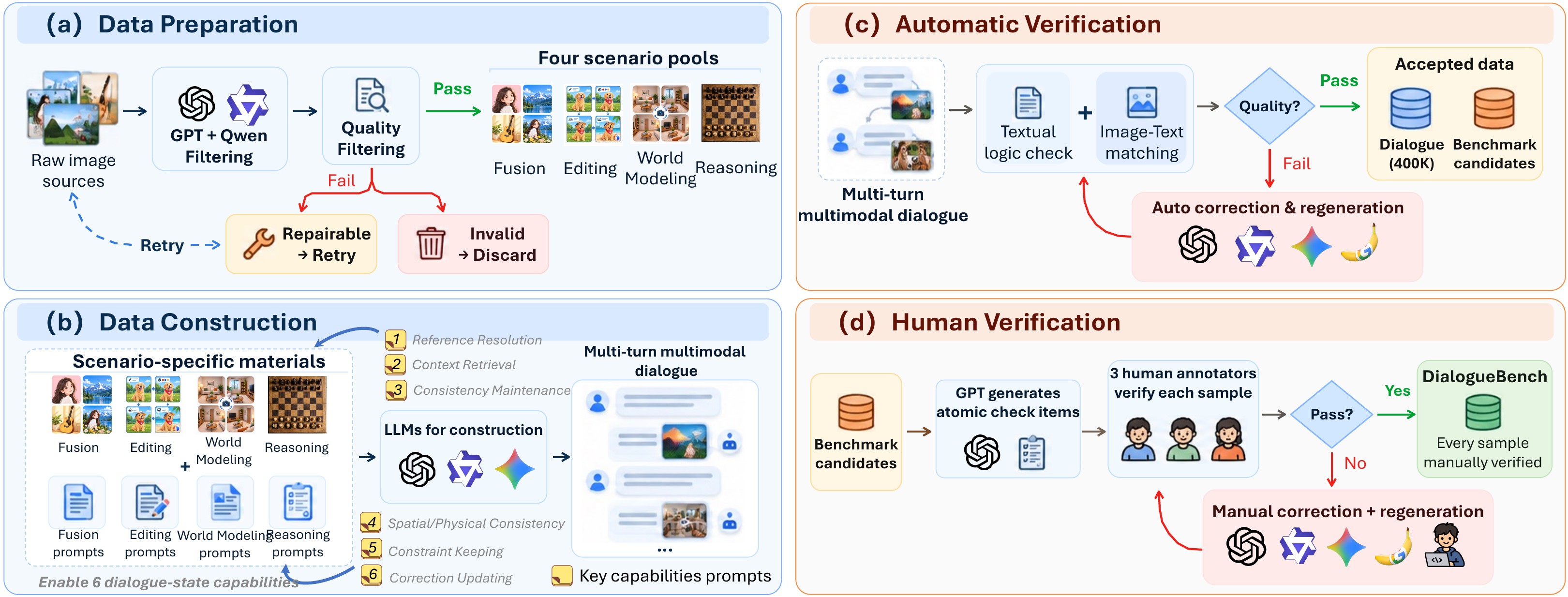}
\caption{\textbf{Overview of the \dialoguename data pipeline.} The pipeline comprises four stages: (a) Data Preparation, (b) Data Construction, (c) Automatic Verification, and (d) Human Verification.}
    \label{fig:pipeline}
    \vspace{-2mm}
\end{figure}

\begin{figure}[t]
\centering
\setlength{\dialoguepanelheight}{0.235\textwidth}
\begin{minipage}[t]{0.22\textwidth}
\centering
\begin{minipage}[c][\dialoguepanelheight][c]{\linewidth}
\centering
\fontsize{6.5}{7.5}\selectfont
\setlength{\tabcolsep}{2pt}
\renewcommand{\arraystretch}{1.0}
\begin{tabular*}{\linewidth}{@{\extracolsep{\fill}}lr@{}}
\toprule
\textbf{Statistic} & \textbf{Number} \\
\midrule

\textbf{Total Dialogues} & 402,639 \\
\hspace{0.6em}$\geq$ 4 Images & 321,868 \\
\hspace{0.6em}$\geq$ 5 Images & 215,027 \\
\hspace{0.6em}$\geq$ 6 Images & 138,669 \\

\addlinespace[2pt]

\textbf{Total Images} & 2,047,236 \\
\hspace{0.6em}Avg. / Dialogue & 5.08 \\

\addlinespace[2pt]

\textbf{Total Turns} & 2,643,291 \\
\hspace{0.6em}Avg. / Dialogue & 6.56 \\
\hspace{0.6em}Max. / Dialogue & 43 \\

\bottomrule
\end{tabular*}
\end{minipage}
\par\vspace{2pt}
\captionof{table}{Key Statistics of our Dataset.}
\label{tab:statistics}
\end{minipage}
\hfill
\begin{minipage}[t]{0.48\textwidth}
\centering
\begin{minipage}[c][1.1\dialoguepanelheight][c]{\linewidth}
\centering
\includegraphics[width=0.9\linewidth]{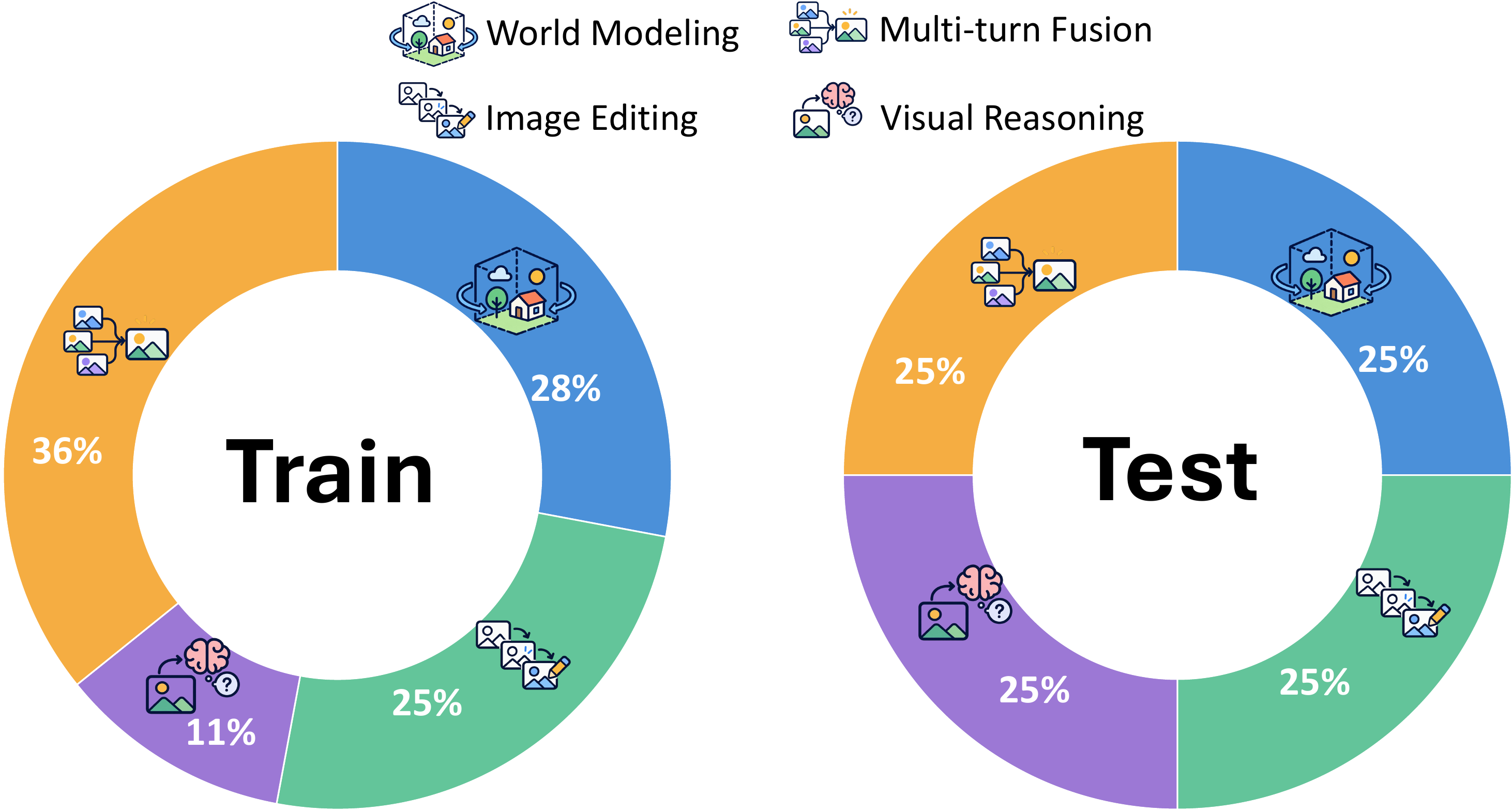}
\end{minipage}
\par\vspace{-1mm}
\captionof{figure}{Distribution of the four interaction scenarios in the training and test sets.}
\label{fig:scenario_distributions}
\end{minipage}
\hfill
\begin{minipage}[t]{0.28\textwidth}
\centering
\begin{minipage}[c][1.\dialoguepanelheight][c]{\linewidth}
\centering
\includegraphics[width=\linewidth]{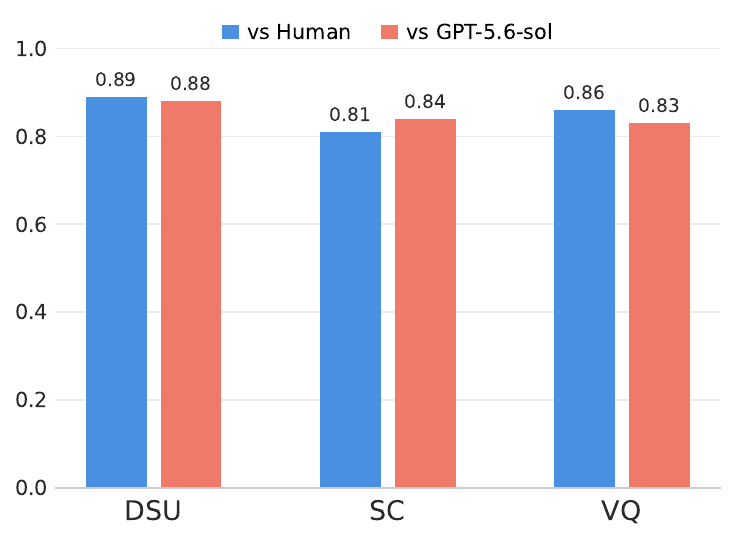}
\end{minipage}
\par\vspace{2pt}
\captionof{figure}{Evaluation reliability of Our VLM-Judge.}
\label{fig:evaluation_reliability}
\end{minipage}
\vspace{-4mm}
\end{figure}

\textbf{\dialoguedataset.} We construct novel stateful dialogues using visual materials drawn from four existing public datasets~\citep{li2025zebra,chow2026weave,wei2026mico,chen2026macro}.
Our pipeline, illustrated in Fig.~\ref{fig:pipeline}, leverages GPT-5.6 Sol~\citep{openai2026gpt56}, Qwen3.7-Plus~\citep{alibaba2026qwen37plus}, Gemini 3.1 Pro~\citep{google2026gemini31pro}, GPT-Image-2~\citep{openai2026gptimage2}, and Nano Banana 2~\citep{google2026nanobanana2} for data filtering, dialogue construction, and quality control.
Specifically, raw images are filtered into four scenario-specific pools (Fig~\ref{fig:pipeline}.a) and combined with scenario- and capability-oriented prompts to construct multi-turn multimodal stateful dialogues (Fig~\ref{fig:pipeline}.b).
The resulting dialogues are automatically verified for logical coherence and image--text consistency, with failed samples iteratively corrected and regenerated.
Dialogues that pass verification form \dialoguedataset (Fig~\ref{fig:pipeline}.c).
The constructed dialogues span four representative scenarios: \textit{1) Multi-image fusion,} where the assistant integrates multiple user-provided or previously generated images according to the dialogue history;
\textit{2) Image editing,} where the assistant progressively edits an image while preserving prior modifications and visual consistency;
\textit{3) World modeling,} where the assistant predicts and renders scene states under specified viewpoint changes or actions; and
\textit{4) Visual reasoning,} where the assistant generates visual outcomes for proposed hypotheses or solutions to support step-by-step reasoning.

To make these interactions genuinely stateful, we incorporate six key state-management capabilities across the four scenarios:
\textit{(i) Coreference resolution,} which grounds context-dependent expressions to the correct textual or visual entities;
\textit{(ii) Dialogue-history retrieval,} which identifies and uses relevant information from previous turns;
\textit{(iii) Cross-turn consistency,} which preserves established visual attributes and prior edits across turns;
\textit{(iv) Spatial and physical consistency,} which maintains plausible geometry, physical relations, and state transitions;
\textit{(v) Constraint adherence,} which consistently follows user-specified requirements and restrictions; and
\textit{(vi) Correction-based state updating,} which incorporates user corrections and replaces outdated information accordingly.
For example, Fig.~\ref{fig:dataset_examples} presents representative dialogues across the four scenarios, illustrating how multimodal dialogue states are progressively established, maintained, and updated through the joint use of these capabilities.
Additional details of the data collection and representative examples are provided in Appendices~\ref{app:data_construction} and~\ref{app:qualitative_examples}, respectively.

\textbf{\dialoguebench.} \dialoguebench is constructed with LLM assistance, followed by exhaustive human verification and correction by annotators with graduate-level training in STEM disciplines~(Fig~\ref{fig:pipeline}.d).
Throughout data construction, we strictly isolate the benchmark from the training dataset to prevent data leakage; as a final safeguard, we compare the SHA-256 hashes of all images across \dialoguebench and \dialoguedataset, confirming that no exact duplicates exist between them.
It comprises 400 multi-turn dialogues covering the four scenarios described above.
The benchmark contains complex instructions that require reasoning over cross-turn context, jointly evaluating models' capabilities in ability to track, maintain, and update multimodal dialogue states and translating the understanding of the state into visual implementation.

\subsection{Evaluation Setting and Metrics}
\label{eval_setting}
To better quantify models' stateful interaction capabilities in unified multi-turn multimodal dialogues, we evaluate them from two complementary perspectives: dialogue state understanding and visual realization.
Specifically, we adopt a VLM-as-a-Judge framework for automatic evaluation.

For DSU, we construct four types of QA pairs for each dialogue: \textit{1) Reference Resolution} measures whether the model can correctly resolve referring expressions in the current instruction and associate them with the corresponding entities in the dialogue history or previous images.
\textit{2) Context Retrieval} measures whether the model can retrieve visual or textual information relevant to the current task from the multi-turn interaction history.
\textit{3) Constraint Tracking} evaluates whether the model can identify and continuously retain constraints introduced in previous turns that remain valid at the current turn;
and \textit{4) Correction Updating} measures whether the model can discard invalidated states and update the multimodal dialogue state after the user corrects previously established information, requirements, or states.

For VR, we evaluate generated or edited visual outputs from two aspects: \textit{Visual Quality}, which measures overall perceptual quality, and \textit{State Consistency}, which uses fine-grained atomic criteria to assess whether the output faithfully reflects all valid multimodal dialogue states accumulated up to the current turn.
For each scenario, we first average the four DSU metrics and then combine the resulting understanding score with State Consistency and Visual Quality to obtain the aggregate score.
For models without understanding capabilities, the aggregate score is computed from State Consistency and Visual Quality alone.
The detailed evaluation templates and score calculation are provided in Appendix~\ref{app:judge_template} and Appendix~\ref{app:score_calculation}, respectively.

\subsection{Data Statistics}
\label{data_statics}

Table~\ref{tab:statistics} summarizes the key dataset statistics.
Most dialogues contain at least four images and have an average of 6.56 conversational turns.
As shown in Fig.~\ref{fig:scenario_distributions}, the training set includes all four interaction scenarios in varying proportions, while the test set is evenly balanced across them to enable controlled cross-scenario evaluation.
Detailed statistics are provided in Appendix~\ref{app:data_statistics}.

\section{Experiment}
\label{Experiment}
\begin{table}[t]
\centering
\caption{\textbf{Results on \dialoguebench across four interaction scenarios.}
\scalebox{0.72}{\tableicon{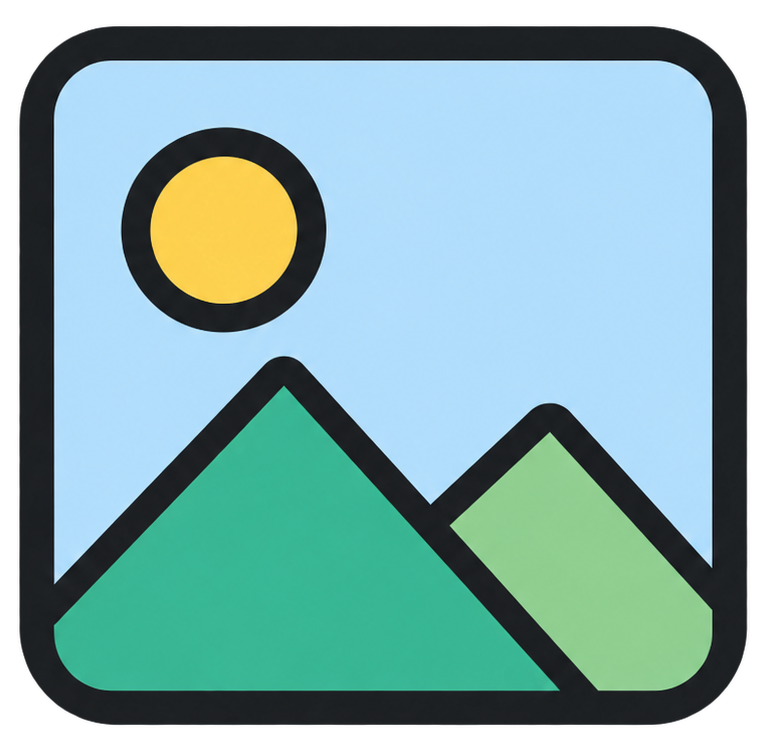}} denotes models supporting image generation/editing only; \scalebox{0.72}{\tableicon{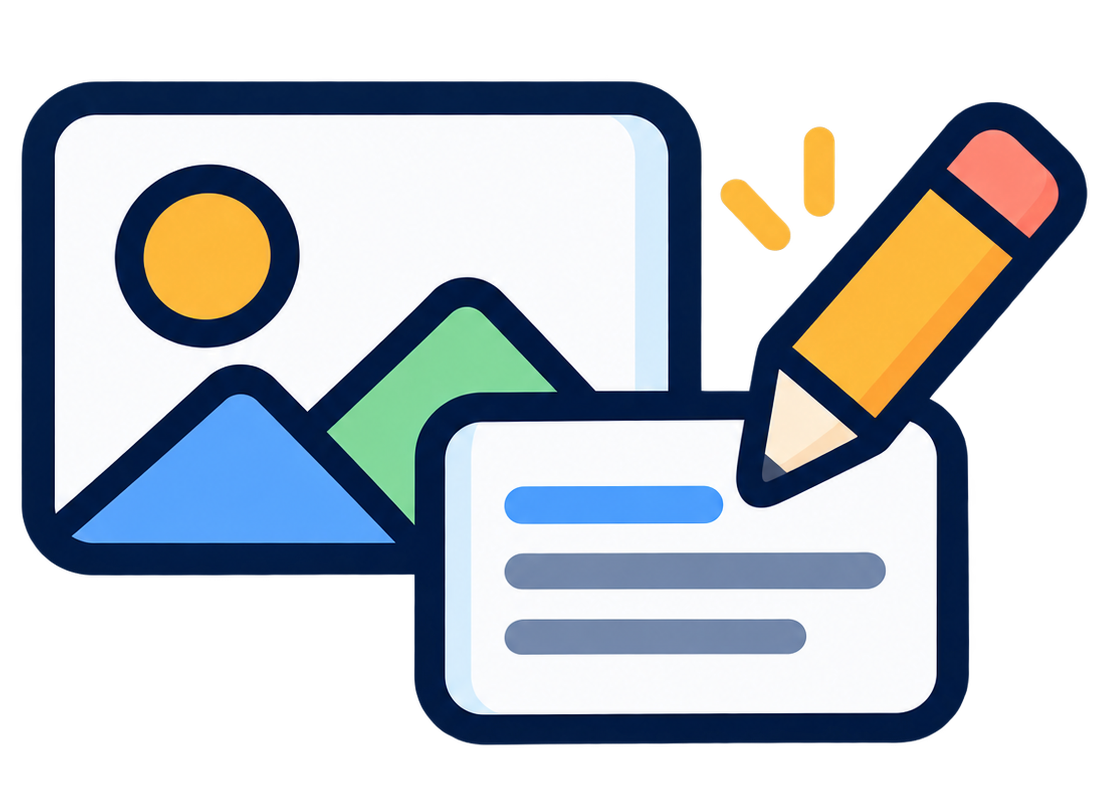}} denotes models supporting both understanding and generation.
\scalebox{0.72}{\tableicon{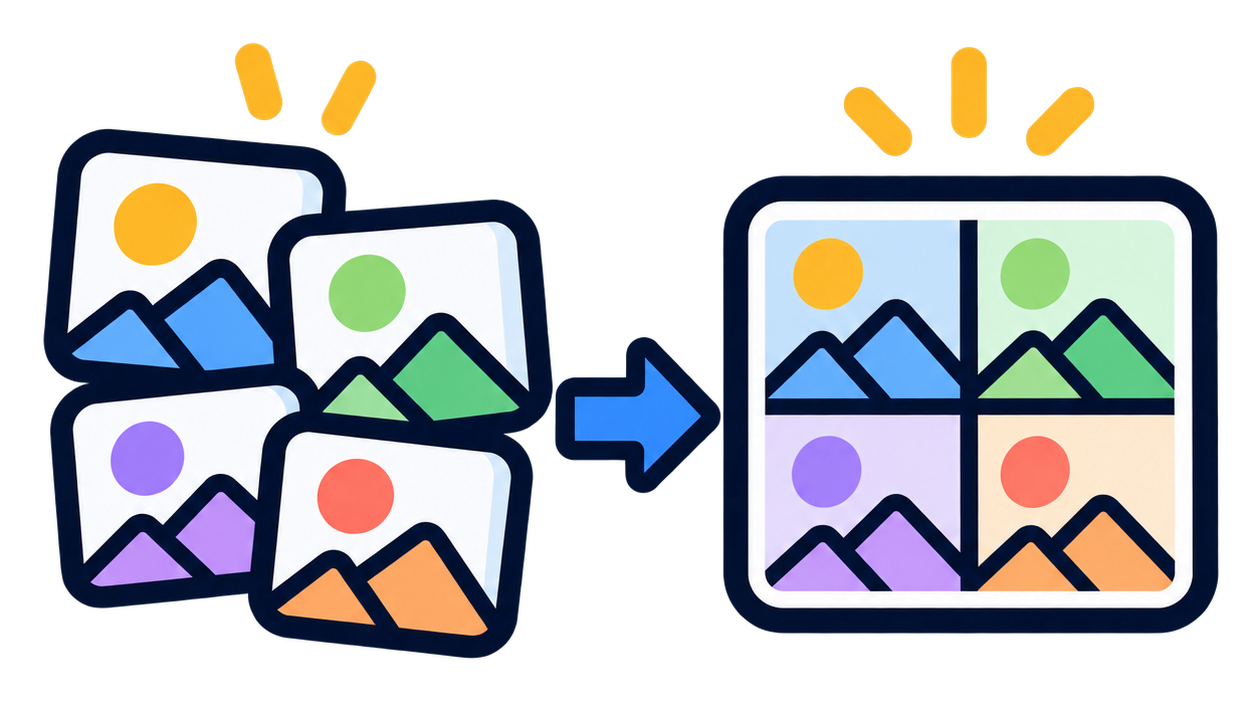}} denotes concatenated image inputs; \scalebox{0.72}{\tableicon{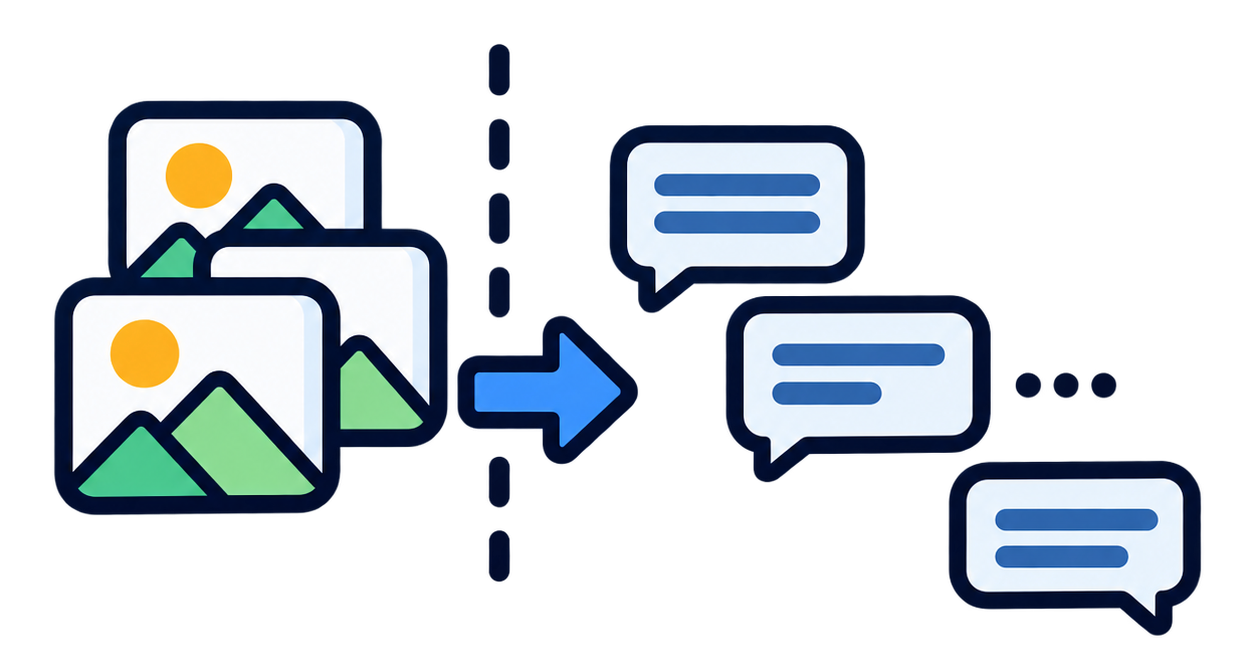}} denotes placing all available images at the beginning; and \scalebox{0.72}{\tableicon{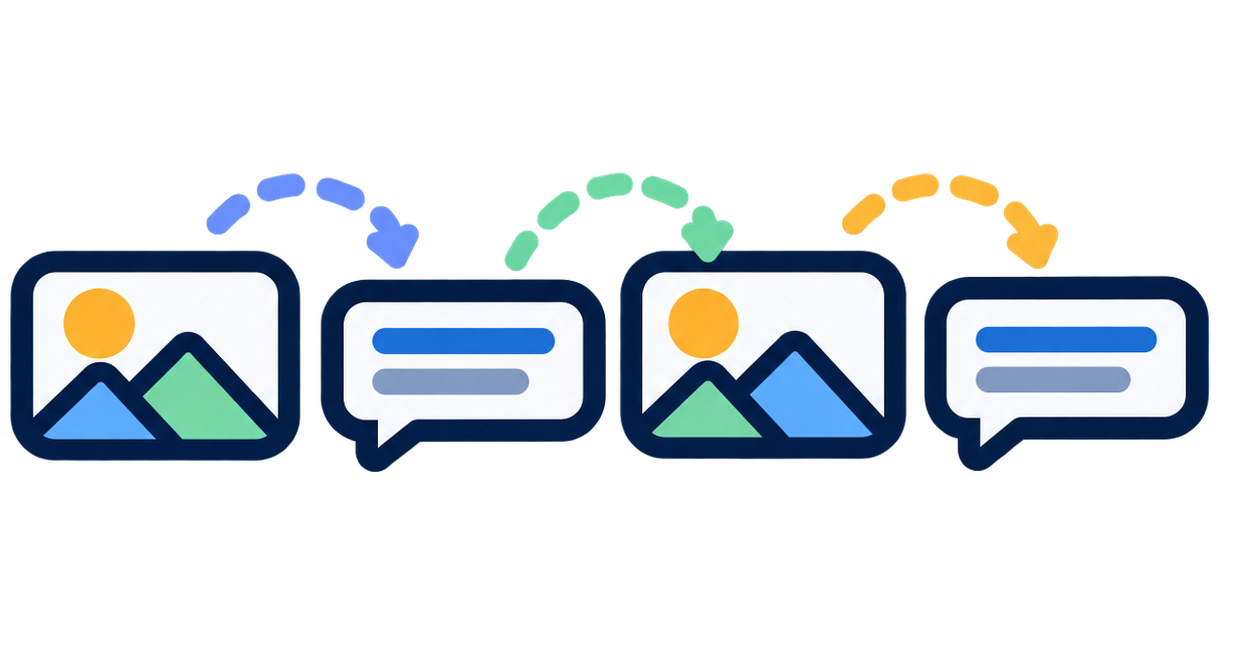}} denotes interleaved image--text inputs.
{\setlength{\fboxsep}{1pt}%
\colorbox{red!12}{Red}, \colorbox{blue!12}{blue}, and \colorbox{yellow!20}{yellow}} cells highlight the best results among closed-source models, open-source image-editing models, and UMMs.
Underlined results denote the second-best performance within each model group.
}
\label{tab:main_results}
\small
\setlength{\tabcolsep}{5pt}
\renewcommand{\arraystretch}{0.95}
\resizebox{\textwidth}{!}{%
\begin{tabular}{@{}clcccccccc@{}}
\toprule
\textbf{Index}
& \textbf{Model}
& \textbf{Size}
& \textbf{Modality}
& \textbf{Format}
& \textbf{Fusion}
& \textbf{Editing}
& \textbf{Modeling}
& \textbf{Reasoning}
& \textbf{Overall} \\
\midrule

1 & GPT-Image-2~\citep{openai2026gptimage2}
& -- & \tableicon{image_only} & \tableicon{image_first} & \cellcolor{red!12}86.78 & \cellcolor{red!12}90.54 & \cellcolor{red!12}81.40 & \underline{73.46} & \cellcolor{red!12}83.05 \\

2 & Nano Banana 2~\citep{google2026nanobanana2}
& -- & \tableicon{image_only} & \tableicon{image_then_text} & \underline{84.73} & \underline{87.32} & \underline{74.74} & \cellcolor{red!12}76.93 & \underline{80.93} \\

\midrule

3 & Step1X-v1p2~\citep{liu2025step1x}
& 12B & \tableicon{image_only} & \tableicon{concat} & 56.08 & 63.36 & 51.66 & 48.00 & 54.77 \\

4 & Qwen-Edit~\citep{wu2025qwen}
& 20B & \tableicon{image_only} & \tableicon{image_then_text} & 49.66 & 54.70 & 28.14 & 37.72 & 42.55 \\

5 & FLUX.2 [dev]~\citep{labs2025flux2}
& 32B & \tableicon{image_only} & \tableicon{concat} & \cellcolor{blue!12}72.44 & \cellcolor{blue!12}74.54 & \cellcolor{blue!12}65.54 & \cellcolor{blue!12}56.68 & \cellcolor{blue!12}67.30 \\

6 & UniPic 3.0~\citep{wei2026skywork}
& 20B & \tableicon{image_only} & \tableicon{concat} & \underline{63.44} & \underline{70.66} & \underline{56.96} & \underline{48.82} & \underline{59.97} \\

\midrule

7 & OmniGen~\citep{xiao2025omnigen}
& 3.8B & \tableicon{image_only} & \tableicon{image_then_text} & 48.24 & 58.12 & 41.82 & 33.94 & 45.53 \\

8 & OmniGen2~\citep{wu2025omnigen2}
& 7B & \tableicon{gen_and_com} & \tableicon{image_then_text} & 50.10 & 51.70 & 55.12 & 47.62 & 51.13 \\

9 & Janus-Pro~\citep{chen2025janus}
& 7B & \tableicon{gen_and_com} & \tableicon{image_then_text} & 45.94 & 41.94 & 42.56 & 31.86 & 40.58 \\

10 & Show-o2~\citep{xie2026show}
& 7B & \tableicon{gen_and_com} & \tableicon{concat} & 40.82 & 45.12 & 41.78 & 29.42 & 39.29 \\

11 & Ovis-U1~\citep{wang2025ovis}
& 3B & \tableicon{gen_and_com} & \tableicon{concat} & 68.72 & 46.24 & 57.72 & 43.42 & 54.03 \\

12 & Emu3.5~\citep{cui2025emu3}
& 34B & \tableicon{gen_and_com} & \tableicon{image_then_text} & \underline{74.48} & 60.10 & \cellcolor{yellow!20}74.06 & 60.98 & \underline{67.41} \\

13 & InternVL-U~\citep{tian2026internvl}
& 4B & \tableicon{gen_and_com} & \tableicon{concat} & 66.72 & 66.12 & 66.48 & 46.52 & 61.46 \\

14 & UniAR~\citep{peng2026unified}
& 10B & \tableicon{gen_and_com} & \tableicon{image_then_text} & 57.06 & 41.82 & 41.80 & 35.92 & 44.15 \\

\cmidrule(l){1-10}

15 & BAGEL~\citep{deng2025emerging}
& 7B & \tableicon{gen_and_com} & \tableicon{image_then_text} & 62.44 & \underline{67.40} & 68.16 & \underline{62.06} & 65.02 \\

 & \textbf{+ \dialoguedataset{}}
& 7B & \tableicon{gen_and_com} & \tableicon{image_then_text} & \cellcolor{yellow!20}78.98\,$\uparrow$ & \cellcolor{yellow!20}84.96\,$\uparrow$ & \underline{73.26}\,$\uparrow$ & \cellcolor{yellow!20}75.80\,$\uparrow$ & \cellcolor{yellow!20}78.25\,$\uparrow$ \\

\bottomrule
\end{tabular}
}
\vspace{0.5mm}

\end{table}

In this section, we first evaluate 15 models on \dialoguebench, revealing substantial limitations of existing models in understanding, maintaining, updating, and visually realizing multimodal dialogue states.
We then validate the effectiveness and high quality of \dialoguedataset{} by fine-tuning BAGEL in Sec.~\ref{train_on_dialogue-400k}.
Finally,  we evaluate the reliability of our VLM-as-a-Judge evaluation framework.

\subsection{\dialoguebench{}}
\label{DialogueBenchmark}
\textbf{Settings.} We evaluate 2 close-source models, 4 open-source image editing models, and 9 unified multimodal models on \dialoguebench, with the results reported in Table~\ref{tab:main_results}.
Complete capability-level results for each interaction scenario are provided in Appendix~\ref{app:full_understanding_results}.
To isolate models' ability to leverage multi-turn dialogue context from error propagation caused by previous generations, we adopt a \textit{GT-History} evaluation setting.
For image placement, we consider three configurations depending on the model setup:
(1) yes-first, where each image is inserted at its first-mentioned position in the dialogue, (2) yes-front, where all images are placed at the beginning of the input, and (3) concatenate, where all images are concatenated into one image for models that cannot directly process sequential multi-image inputs.
Implementation details for the history settings and input organizations are provided in Appendix~\ref{app:history_input_format}.
Based on the results, we can derive the following key findings:

\textbf{Stateful Multi-Turn Multimodal Interaction Remains Challenging.}
As shown in Table~\ref{tab:main_results}, the best average scores achieved by existing UMMs and open-source image editing models on YANAMI-Bench are only 67.41 and 67.30, respectively, indicating that current models still face substantial challenges in multimodal multi-turn stateful interaction.
Further examination across different interaction scenarios reveals clear performance variation, with Visual Reasoning yielding the lowest scores for most evaluated models.
In contrast, GPT-Image-2 and Nano Banana 2 achieve average scores of 83.05 and 80.93, respectively, outperforming current open-source models overall and suggesting that a noticeable gap remains between open- and closed-source models in multimodal multi-turn stateful interaction.

\noindent\textbf{Multi-turn Dialogue-State Understanding Remains Challenging.}\par
\begingroup
\setlength{\columnsep}{9pt}
\setlength{\intextsep}{6pt}
\begin{table}[!htbp]
\centering
\captionsetup{skip=7pt}
\caption{Selected average DSU results. An upward arrow indicates an improvement over BAGEL. Full results are provided in Appendix~\ref{app:full_understanding_results}.}
\label{tab:state_understanding}
\small
\setlength{\tabcolsep}{6pt}
\renewcommand{\arraystretch}{1.15}
\begin{tabularx}{\linewidth}{l*{5}{>{\centering\arraybackslash}X}}
\toprule
\textbf{Model}
& \textbf{Fus.}
& \textbf{Edi.}
& \textbf{Mod.}
& \textbf{Rea.}
& \textbf{Avg.} \\
\midrule
Show-o2 & 74.00 & 64.25 & 79.50 & 54.00 & 67.94 \\
BAGEL & 85.75 & 76.50 & 85.25 & 78.25 & 81.44 \\
\rowcolor{yellow!10}
\textbf{BAGEL + Ours} & \textbf{96.25}\,$\uparrow$ & \textbf{98.00}\,$\uparrow$ & \textbf{95.50}\,$\uparrow$ & \textbf{93.00}\,$\uparrow$ & \textbf{95.69}\,$\uparrow$ \\
\bottomrule
\end{tabularx}
\end{table}

We further find that models struggle not only with generating and editing visual outputs, but also with understanding the current multimodal dialogue state.
As shown in Table~\ref{tab:state_understanding}, the average DSU scores of most UMMs remain below 82, indicating that they still have difficulty accurately recovering and maintaining the currently valid dialogue state as it evolves over multiple turns, including historical references, relevant contextual information, persistent constraints, and state updates following user corrections.
\par
\endgroup
\begin{figure}[t]
    \centering
    \begin{subfigure}[t]{0.29\textwidth}
        \centering
        \includegraphics[width=\linewidth]{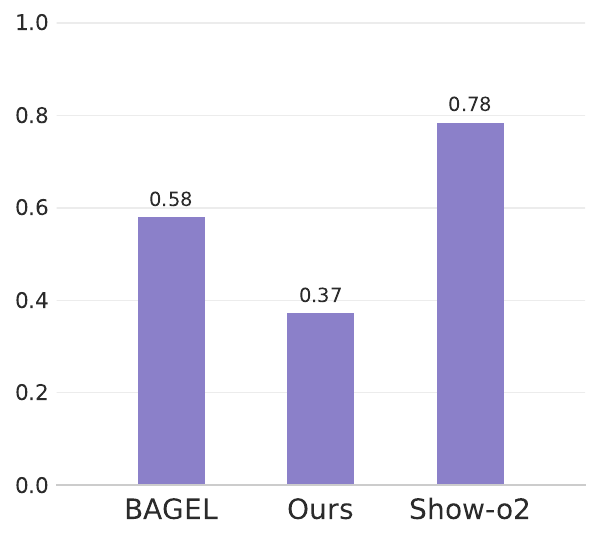}
        \caption[Understanding--Realization Gap]{Und. Realization Gap}
        \label{fig:understanding_realization_gap}
    \end{subfigure}
        \begin{subfigure}[t]{0.37\textwidth}
        \centering
        \includegraphics[width=\linewidth]{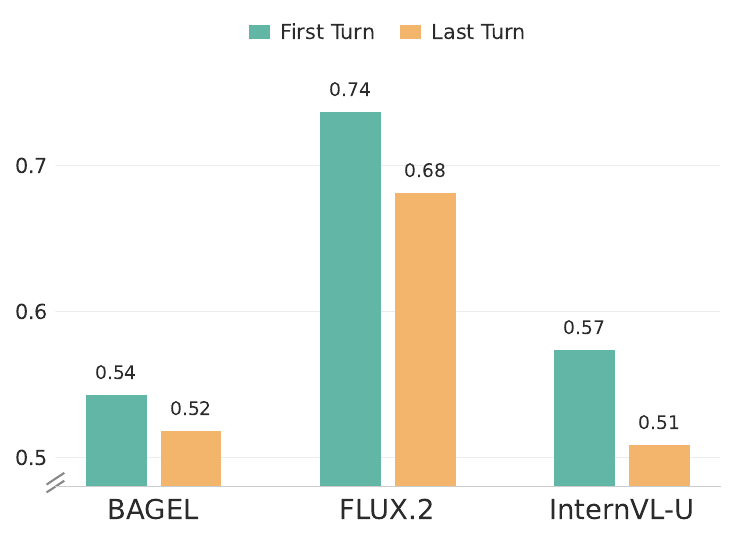}
        \caption[First vs.\ Last Generation Turn]{First vs.\ Last Generation Turn}
        \label{fig:first_last_turn}
    \end{subfigure}
    \begin{subfigure}[t]{0.30\textwidth}
        \centering
        \includegraphics[width=\linewidth]{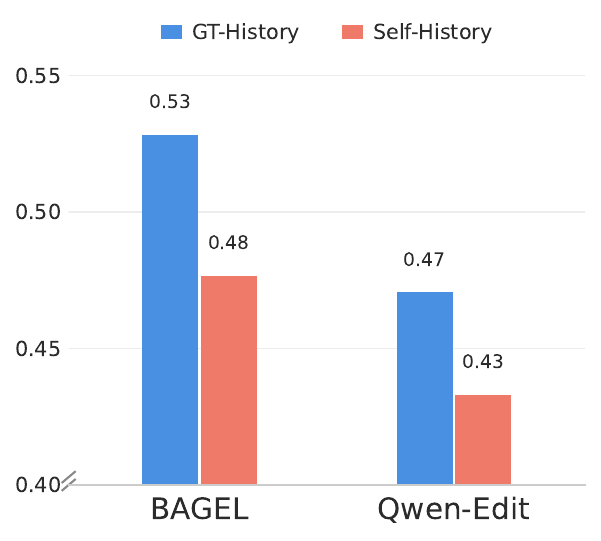}
        \caption[GT-History vs.\ Self-History]{GT-History vs.\ Self-History}
        \label{fig:gt_self_history}
    \end{subfigure}

    \caption{\textbf{Analysis of stateful multi-turn multimodal interaction.}
    (a) Measuring the understanding-realization gap, lower is better;
    (b) comparing performance at the first and last image-generation turns under GT-History, to assess the effect of increasingly accumulated context and state dependencies;
    (c) comparing GT-History with Self-History on dialogues to measure error accumulation when self-generated images are used as subsequent visual context.
}\label{fig:stateful_interaction_analysis}
\end{figure}
\Needspace{5\baselineskip}
\textbf{Dialogue-State Understanding–Visual Realization Gap.}
We further investigate the gap between state understanding and visual realization in multimodal multi-turn interaction.
For a fair comparison, we retain only state-understanding items on which all evaluated models receive full credit and compute each model's normalized realization score using the same aligned generation-side capability checks; the understanding--realization gap is then defined as one minus this conditional realization score.
Details of the evaluation protocol are provided in Appendix~\ref{app:gap_evaluation}.
As shown in Fig.~\ref{fig:understanding_realization_gap}, even when a model correctly understands a currently valid constraint or a corrected state, it does not always faithfully realize that information in the generated output.
This reveals a clear gap between state understanding and visual realization in existing UMMs.

\textbf{Performance Across Dialogue Progression.}
We analyze how model performance changes as multi-turn interactions progress.
Fig.~\ref{fig:first_last_turn} compares the performance of three representative models at the first and last image-generation turns of dialogues containing at least two generation turns.
Since our evaluation adopts the GT-History setting, errors from previous generations do not propagate to later turns.
This comparison therefore more directly reflects how models handle increasingly accumulated context and state dependencies over long-horizon interactions.
The results show performance drops of 2\%, 6\%, and 6\% for BAGEL, FLUX.2, and InternVL-U, respectively, suggesting that models still struggle to stably maintain and realize the complete multimodal dialogue state as dialogue length and state complexity increase.

\textbf{GT-History vs. Self-History.}
To better reflect realistic multi-turn interaction, we additionally evaluate a \textit{Self-History} setting, in which images generated by the model in previous turns are used as visual context for subsequent turns.
To focus on interactions in which generated visual history can affect subsequent outputs, we include only dialogues requiring more than one generation turn.
As shown in Fig.~\ref{fig:gt_self_history}, performance under Self-History decreases by 5.16\% and 3.77\% in BAGEL and Qwen-Edit, respectively, compared with GT-History.
This result suggests that, although models can perform reasonably well when conditioned on correct visual histories, error accumulation from self-generated histories can substantially degrade stateful interaction performance during realistic multi-turn rollouts.
More results and detailed breakdowns are provided in Appendix~\ref{app:complete_experimental_results}.

\subsection{Train on \dialoguedataset{}}
\label{train_on_dialogue-400k}
To validate the effectiveness of \dialoguedataset{}, we fine-tune BAGEL on our dataset, with detailed training configurations provided in Appendix~\ref{app:training_details}.
The results demonstrate that \dialoguedataset{} substantially improves image generation, image editing, and multi-turn multimodal dialogue state modeling, while largely preserving the model's general visual understanding capabilities.
\begin{table}[t]
\centering
\caption{\textbf{Comparison of different models on visual understanding tasks and GenEval.} An upward arrow indicates an improvement over BAGEL.}
\label{tab:understanding_geneval}
\small
\setlength{\tabcolsep}{4pt}
\renewcommand{\arraystretch}{0.95}

\resizebox{\linewidth}{!}{%
\begin{tabular}{l|ccc|ccccccc}
\toprule
\multirow{2}{*}{\textbf{Model}}
& \multicolumn{3}{c|}{\textbf{Understanding}}
& \multicolumn{7}{c}{\textbf{GenEval}} \\
\cmidrule(lr){2-4}
\cmidrule(lr){5-11}

& \textbf{POPE}
& \textbf{MME-P}
& \textbf{MMBench}
& \textbf{Single Obj.}
& \textbf{Two Obj.}
& \textbf{Count.}
& \textbf{Color}
& \textbf{Position}
& \textbf{Attri.}
& \textbf{Overall} \\
\midrule

Emu3~\citep{wang2024emu3}
& 85.2
& 1243.8
& 58.5
& 0.99
& 0.81
& 0.42
& 0.80
& 0.49
& 0.45
& 0.66 \\

Show-o~\citep{xie2025show}
& 80.0
& 1097.2
& --
& 0.98
& 0.80
& 0.66
& 0.84
& 0.31
& 0.50
& 0.68 \\

Janus-Pro-7B~\citep{chen2025janus}
& 87.4
& 1567.1
& 79.2
& 0.99
& 0.89
& 0.59
& 0.90
& 0.79
& 0.66
& 0.80 \\

MetaQuery-XL~\citep{pan2025transfer}
& --
& 1685.2
& 83.5
& --
& --
& --
& --
& --
& --
& 0.80 \\

Ovis-U1~\citep{wang2025ovis}
& --
& 1508.0
& 77.8
& 0.98
& 0.98
& 0.90
& 0.92
& 0.79
& 0.75
& 0.89 \\

Show-o2~\citep{xie2026show}
& --
& 1620.5
& 79.3
& 1.00
& 0.87
& 0.58
& 0.92
& 0.52
& 0.62
& 0.76 \\

BLIP3-o-8B~\citep{chen2025blip3}
& --
& 1682.6
& 83.5
& 1.00
& 0.92
& 0.63
& 0.91
& 0.86
& 0.67
& 0.83 \\

InternVL-U~\citep{tian2026internvl}
& --
& 1607.5
& 80.5
& 0.99
& 0.94
& 0.74
& 0.91
& 0.77
& 0.74
& 0.85 \\

Emu3.5~\citep{cui2025emu3}
& --
& 781.1
& --
& --
& --
& --
& --
& --& --
& 0.86 \\

UniAR~\citep{peng2026unified}
& --
& --
& --
& 1.00
& 0.95
& 0.75
& 0.94
& 0.77
& 0.67
& 0.85 \\

OmniGen2~\citep{wu2025omnigen2}
& --
& 1584.4
& 79.1
& 1.00
& 0.95
& 0.64
& 0.88
& 0.55
& 0.76
& 0.80 \\

\midrule

BAGEL~\citep{deng2025emerging}
& 86.5
& 1647.9
& 85.0
& 1.00
& 0.89
& 0.70
& 0.82
& 0.47
& 0.49
& 0.73 \\

\rowcolor{yellow!10}
\textbf{+ \dialoguedataset{}}
& 87.4\,$\uparrow$
& 1616.9
& 89.1\,$\uparrow$
& 0.99
& 0.90\,$\uparrow$
& 0.73\,$\uparrow$
& 0.81
& 0.52\,$\uparrow$
& 0.55\,$\uparrow$
& 0.75\,$\uparrow$\\

\bottomrule
\end{tabular}%
}

\end{table}

Fine-tuned BAGEL performances on YANAMI-Bench across four interaction scenarios:
\textit{(i) Visual Understanding.}
The fine-tuned BAGEL largely preserves its original general visual understanding capabilities.
As shown in Table~\ref{tab:understanding_geneval}, the model achieves improvements of 4.1\% and 0.9\% on MMBench and POPE, respectively, while exhibiting only a slight decrease on MME-P, indicating that training on \dialoguedataset{}  preserves general visual understanding capability.
\textit{(ii) Image Generation.}
As shown in Table~\ref{tab:understanding_geneval}, the fine-tuned BAGEL improves the overall score on GenEval by 2\%, demonstrating that \dialoguedataset{} enhances the model's general image generation capability.
\textit{(iii) Image Editing.}
\begin{table}[t]
\centering
\caption{\textbf{Comparison of fine-tuned BAGEL and other models on GEdit-EN-full benchmark.} An upward arrow indicates an improvement over BAGEL.}
\label{tab:gedit}
\small
\setlength{\tabcolsep}{3pt}
\renewcommand{\arraystretch}{0.95}
\resizebox{\textwidth}{!}{
\begin{tabular}{@{}lccccccccccc@{\hspace{3pt}}c@{}}
\toprule
\textbf{Model}
& \textbf{BG}
& \textbf{Color}
& \textbf{Mat.}
& \textbf{Motion}
& \textbf{Port.}
& \textbf{Style}
& \textbf{Add}
& \textbf{Remove}
& \textbf{Replace}
& \textbf{Text}
& \textbf{Tone}
& \textbf{Avg.} \\
\midrule

AnyEdit~\citep{yu2025anyedit}
& 4.31 & 4.25 & 2.64 & 0.67 & 1.90 & 1.95
& 3.72 & 3.75 & 3.23 & 0.77 & 4.21 & 2.85 \\

InstructPix2Pix~\citep{brooks2023instructpix2pix}
& 3.94 & 5.40 & 3.52 & 1.27 & 2.62 & 4.39
& 3.07 & 1.50 & 3.48 & 1.13 & 5.10 & 3.22 \\

OmniGen~\citep{xiao2025omnigen}
& 5.23 & 5.93 & 5.44 & 3.12 & 3.17 & 4.88
& 6.33 & 6.35 & 5.34 & 4.31 & 4.96 & 5.01 \\

Step1X-Edit~\citep{liu2025step1x}
& 7.03 & 6.26 & 6.46 & 3.66 & 5.23 & 7.24
& 7.17 & 6.42 & 7.39 & 7.40 & 6.62 & 6.44 \\

Step1X-Edit-v1p2~\citep{liu2025step1x}
& -- & -- & -- & -- & -- & --
& -- & -- & -- & -- & -- & 7.24 \\

OmniGen2~\citep{wu2025omnigen2}
& 6.99 & 6.66 & 4.88 & 2.55 & 3.66 & 6.08
& 7.09 & 6.60 & 6.65 & 4.49 & 6.03 & 5.57 \\

EditMGT~\citep{chow2026editmgt}
& 7.69 & 7.71 & 5.77 & 3.84 & 5.13 & 6.53
& 6.13 & 5.24 & 5.56 & 4.53 & 6.42 & 5.87 \\

GoT-6B~\citep{fang2025got}
& 4.11 & 5.75 & 3.04 & 1.71 & 2.69 & 4.72
& 5.77 & 4.59 & 5.65 & 1.16 & 4.24 & 3.95 \\

VAREdit-8B~\citep{mao2026visual}
& 6.77 & 6.64 & 5.40 & 3.33 & 4.20 & 6.46
& 5.86 & 7.29 & 6.67 & 3.87 & 6.54 & 5.73 \\

FLUX.1 Kontext [dev]~\citep{labs2025flux}
& 7.06 & 7.03 & 5.52 & 5.62 & 4.68 & 5.55
& 6.95 & 6.76 & 6.13 & 6.10 & 7.48 & 6.26 \\

FLUX.2 [dev]~\citep{labs2025flux2}
& -- & -- & -- & -- & -- & --
& -- & -- & -- & -- & -- & 7.41 \\

Qwen-Image-Edit~\citep{wu2025qwen}
& -- & -- & -- & -- & -- & --
& -- & -- & -- & -- & -- & 7.88 \\

Ovis-U1~\citep{wang2025ovis}
& 7.49 & 6.88 & 6.21 & 4.79 & 5.98 & 6.46
& 7.49 & 7.25 & 7.27 & 4.48 & 6.31 & 6.42 \\
\midrule
BAGEL~\citep{deng2025emerging}
& 7.44 & 6.99 & 6.26 & 5.09 & 4.82 & 6.04
& 7.94 & 7.37 & 7.31 & 7.16 & 6.17 & 6.52 \\
\rowcolor{yellow!10}
\textbf{+ \dialoguedataset{}}
& 7.58\,$\uparrow$ & 7.76\,$\uparrow$ & 6.42\,$\uparrow$ & 7.06\,$\uparrow$ & 6.11\,$\uparrow$ & 6.20\,$\uparrow$
& 8.44\,$\uparrow$ & 7.74\,$\uparrow$ & 7.55\,$\uparrow$ & 7.11 & 7.08\,$\uparrow$ & 7.19\,$\uparrow$ \\

\bottomrule
\end{tabular}
}

\end{table}

As shown in Table~\ref{tab:gedit}, the fine-tuned BAGEL improves the overall score on GEdit-Bench-EN by 10.28\%, further demonstrating the effectiveness of \dialoguedataset{} in enhancing image editing capability.
\textit{(iv) Multimodal Dialogue Stateful Interaction.}
As shown in Table~\ref{tab:main_results}, the fine-tuned model improves by 13.23\% over the original BAGEL on \dialoguebench{}.
This result demonstrates that \dialoguedataset{} not only improves general image generation and editing capabilities, but also substantially enhances the model's ability to understand, maintain, update, and visually realize multimodal dialogue states throughout multi-turn interactions.
\subsection{Reliability of Judge Usage}
\label{Reliability_of_Judge_Usage}

To assess the reliability of our VLM-as-a-Judge evaluation framework, we conduct a human evaluation study involving four expert annotators.
Specifically, we sample 100 instances from each of three representative models: BAGEL+\dialoguename, Emu3.5, and Nano Banana 2, and collect human ratings for comparison.
We compute Pearson correlation coefficients\citep{benesty2009pearson} between the scores produced by Qwen-3.8-Flash-Next\citep{qwen3.8flashnext} and the aggregated human ratings, and further compare its evaluations with those produced by GPT-5.6-sol, as shown in Fig.~\ref{fig:evaluation_reliability}.

The results show that the correlations between Qwen-3.8-Flash-Next and human ratings consistently exceed 0.80, indicating strong agreement between the VLM judge and human evaluation.
Moreover, the scores produced by Qwen-3.8-Flash-Next and GPT-5.6-sol exhibit high cross-VLM correlation, demonstrating strong consistency across different VLM evaluators.
These results suggest that our evaluation framework aligns well with human judgments and is robust to the choice of VLM judge.

\FloatBarrier
\section{Conclusion}
This work introduces \textbf{\dialoguename{}}, the first large-scale dataset and benchmark for multimodal stateful interaction across understanding, generation, and editing.
\dialoguedataset{} provides 400K multi-turn dialogues, more than 2M images, and 2.6M dialogue turns across four scenarios and six types of contextual dependency.
\dialoguebench{} comprises 400 multi-turn dialogues and jointly evaluates dialogue-state understanding and visual realization.
Experiments show that training on \dialoguedataset{} improves multi-turn contextual understanding, image generation, and image editing while preserving general visual understanding.
Evaluations on \dialoguebench{} further show that existing UMMs still struggle to maintain, update, and visually realize dialogue states.
\dialoguename{} provides a unified foundation for robust multimodal stateful interaction.

\subsection*{AI Use Statement}
LLM-based tools assisted with language polishing, code development, data filtering, dialogue construction, and quality control.
As detailed in Appendix~\ref{app:data_construction}, GPT-5.6 Sol, Qwen3.7-Plus, and Gemini 3.1 Pro were used in the data-construction pipeline, while GPT-Image-2 and Nano Banana 2 supported image generation and repair; Qwen-3.8-Flash-Next was additionally used for automated evaluation under the protocols in Appendix~\ref{app:evaluation}.
All benchmark samples and annotations were manually verified; the authors reviewed AI-assisted writing and code, validated the resulting procedures, and take responsibility for the final content and claims of this work.

\subsection*{Ethics Statement}
\dialoguename{} is intended for research on multimodal stateful interaction and is constructed from existing public datasets and model-generated content.
We follow the applicable licenses and usage conditions of the source datasets and models.
Automated filtering and human verification are applied to improve data quality.
We encourage the research community to use and further develop these resources responsibly.

\subsection*{Reproducibility Statement}
We provide the information needed to reproduce our data construction, training, and evaluation procedures throughout the paper and appendix.
The construction pipeline, quality-control process, benchmark annotation procedure, and detailed data statistics are documented in Appendices~\ref{app:data_construction} and~\ref{app:data_statistics}.
The evaluation protocols, score calculation, history and input formatting, and model-specific inference configurations are described in Appendix~\ref{app:evaluation}, while the fine-tuning configuration is reported in Appendix~\ref{app:training_details}.
Complete experimental results and the verbatim VLM-as-a-Judge prompts are provided in Appendices~\ref{app:complete_experimental_results} and~\ref{app:judge_template}, respectively.
To further support reproducibility, we will release \dialoguedataset{} and \dialoguebench{} on Hugging Face and publish the prompts used for data construction and evaluation in a public GitHub repository.

\clearpage
\bibliography{references}

\clearpage
\appendix
\section*{Appendix}

\makeatletter
\begingroup
  \setcounter{tocdepth}{2}
  \@starttoc{apc}
\endgroup

\newcommand{\addtoappendixcontents}[2]{%
  \addtocontents{apc}{%
    \protect\contentsline{#1}{#2}{\thepage}{\@currentHref}%
    \protected@file@percent
  }%
}

\let\appendixoriginalsection\section
\RenewDocumentCommand{\section}{s o m}{%
  \IfBooleanTF{#1}{%
    \appendixoriginalsection*{#3}%
  }{%
    \IfNoValueTF{#2}{%
      \appendixoriginalsection{#3}%
      \addtoappendixcontents{section}{\protect\numberline{\thesection}#3}%
    }{%
      \appendixoriginalsection[#2]{#3}%
      \addtoappendixcontents{section}{\protect\numberline{\thesection}#2}%
    }%
  }%
}

\let\appendixoriginalsubsection\subsection
\RenewDocumentCommand{\subsection}{s o m}{%
  \IfBooleanTF{#1}{%
    \appendixoriginalsubsection*{#3}%
  }{%
    \IfNoValueTF{#2}{%
      \appendixoriginalsubsection{#3}%
      \addtoappendixcontents{subsection}{\protect\numberline{\thesubsection}#3}%
    }{%
      \appendixoriginalsubsection[#2]{#3}%
      \addtoappendixcontents{subsection}{\protect\numberline{\thesubsection}#2}%
    }%
  }%
}
\makeatother

\newpage
\section{Data Construction Details}
\label{app:data_construction}

\subsection{Comparison with Existing Multimodal Datasets}
\label{app:dataset_comparison}
Table~\ref{tab:dataset_comparison} compares \dialoguedataset{} with representative multimodal understanding, generation, editing, and dialogue datasets.
The comparison separates multi-turn formatting from explicit dialogue-state management: a dataset may be organized as a sequence of turns while still allowing individual target turns to be reformulated as conventional single-turn tasks once the relevant historical content is directly supplied.
\begin{table}[!htbp]
\centering
\caption{\textbf{Comparison with representative multimodal datasets.}
U, G, and E denote visual understanding, image generation, and image editing, respectively.
Multi-turn Format indicates that data are organized as multi-turn user--assistant interactions; Hybrid Evaluation indicates a unified protocol combining complementary evaluation signals rather than merely reporting multiple independent metrics.
Explicit Dialogue-State Management is marked Yes only when a dataset explicitly supervises the establishment, maintenance, invalidation, or update of the currently valid multimodal state; generic history conditioning, context tracking, or a claim of ``stateful dialogue'' alone does not qualify.
Irreducible to Single-Turn indicates that target turns cannot be reformulated as conventional single-turn tasks simply by directly supplying the relevant historical images.
Partial denotes implicit or subset-level coverage.
Scale follows the unit reported by each work and is therefore not directly comparable across all rows.
\textsuperscript{\textdaggerdbl} ChatUMM reports approximately 40B consumed training tokens but does not report the number of dialogues, images, or samples.
\textsuperscript{\S} No official ChatUMM training-data release was publicly available as of 2026-09-24; its categorization is therefore based on the construction procedure and examples reported in the paper. Although ChatUMM describes its interactions as stateful, its published pipeline primarily converts standard single-turn tasks into dialogues through historical dependencies and distractor turns, without explicit dialogue-state supervision. Once the relevant historical input is supplied directly, the reported target turns can generally be reformulated as conventional single-turn generation, editing, or composition tasks.}
\label{tab:dataset_comparison}
\small
\setlength{\tabcolsep}{4.5pt}
\renewcommand{\arraystretch}{1.08}
\resizebox{\textwidth}{!}{%
\begin{tabular}{@{}llcccccc@{}}
\toprule
\textbf{Dataset}
& \textbf{Scale}
& \textbf{Domain}
& \shortstack{\textbf{Open}\\\textbf{Data}}
& \shortstack{\textbf{Multi-turn}\\\textbf{Format}}
& \shortstack{\textbf{Hybrid}\\\textbf{Evaluation}}
& \shortstack{\textbf{Explicit}\\\textbf{Dialogue-State}\\\textbf{Management}}
& \shortstack{\textbf{Irreducible to}\\\textbf{Single-Turn}} \\
\midrule
InstructPix2Pix~\citep{brooks2023instructpix2pix}
& $\sim$450K editing pairs & E & Yes & No & No & No & No \\
AnyEdit~\citep{yu2025anyedit}
& 2.5M editing pairs & E & Yes & No & No & No & No \\
CoMM~\citep{chen2025comm}
& 227K documents & U/G & Yes & No & No & No & No \\
Zebra-CoT~\citep{li2025zebra}
& 182,384 samples & U/G & Yes & No & No & No & No \\
\midrule
Pico-Banana-400K~\citep{qian2026pico}
& 400K editing pairs; 72K multi-turn & E & Yes & Partial & No & No & No \\
WEAVE-100K~\citep{chow2026weave}
& 100,750 dialogues & U/G/E & Yes & Yes & Yes & No & No \\
ChatUMM~\citep{dai2026chatumm}
& Not reported\textsuperscript{\textdaggerdbl} & U/G/E & No\textsuperscript{\S} & Yes & No & No & Partial \\
\midrule
\rowcolor{blue!7}
\textbf{\dialoguedataset{} (Ours)}
& \textbf{402,639 dialogues} & \textbf{U/G/E} & \textbf{Yes} & \textbf{Yes} & \textbf{Yes} & \textbf{Yes} & \textbf{Yes} \\
\bottomrule
\end{tabular}%
}

\end{table}

\subsection{Data Preparation and Composition}
\label{app:data_sources}
We draw raw images from four existing public datasets~\citep{li2025zebra,chow2026weave,wei2026mico,chen2026macro} and process them in the Data Preparation stage shown in Fig.~\ref{fig:pipeline}.a.
Our pipeline leverages GPT-5.6 Sol~\citep{openai2026gpt56}, Qwen3.7-Plus~\citep{alibaba2026qwen37plus}, Gemini 3.1 Pro~\citep{google2026gemini31pro}, GPT-Image-2~\citep{openai2026gptimage2}, and Nano Banana 2~\citep{google2026nanobanana2} for data filtering, dialogue construction, and quality control.
The raw images are first filtered for quality and suitability and then organized into four scenario-specific pools for Multi-Image Fusion, Image Editing, World Modeling, and Visual Reasoning.
For this initial quality-control stage, we use large multimodal models with the task-specific prompt reproduced in Appendix~\ref{app:image_quality_filter_prompt}.
The filter deliberately performs a permissive coarse screening: an image is marked as rejected only when severe corruption, generation failure, invalid content, or extremely poor visibility makes it clearly unusable.
We apply the filter independently to every image and record the resulting quality labels at the image level.
For each image group, if more than half of its images are marked as rejected, we discard the entire group; otherwise, we repair only the rejected images and send the repaired versions through the same filtering process again.
Samples that remain invalid after this process are discarded.
This procedure supplies diverse visual materials for constructing multi-turn multimodal stateful dialogues involving understanding, generation, and editing, without imposing a one-to-one correspondence between a source dataset and an interaction scenario.

\subsubsection{Image Quality Filtering Prompt}
\phantomsection
\label{app:image_quality_filter_prompt}

\begin{tcolorbox}[
  enhanced,
  breakable,
  title={Prompt for Initial Image Quality Filtering},
  fonttitle=\bfseries\small,
  coltitle=black,
  colback=black!1.5,
  colbacktitle=black!7,
  colframe=black!35,
  boxrule=0.5pt,
  arc=1mm,
  left=1.5mm,
  right=1.5mm,
  top=1.2mm,
  bottom=1.2mm,
  before skip=4pt,
  after skip=8pt
]
\begin{Verbatim}
You are an image quality filter responsible for performing a **coarse initial screening** of images.

Your goal is NOT to evaluate aesthetics, fine details, instruction following, or professional image quality.
You should only reject images that are **clearly unusable or extremely low quality**.

Evaluate the input image and determine whether it should be **KEPT** or **REJECTED**.

### REJECT the image only if one or more of the following severe problems occur:

1. **Unrecognizable content**: the image is so corrupted, blurred, distorted, or noisy that its main content cannot be reasonably identified.
2. **Severe generation failure**: the image contains large-scale visual corruption, completely broken structures, extreme artifacts, or incoherent content that makes the image unusable.
3. **Nearly empty or invalid image**: the image is blank, almost entirely a single color, severely cropped to the point that meaningful content is absent, or otherwise appears invalid.
4. **Extremely poor visibility**: the image is so dark, overexposed, low-resolution, or degraded that the main subject or scene cannot be meaningfully perceived.

### KEEP the image if:

- The main subject and scene are recognizable.
- The image contains understandable visual content.
- Imperfections are minor or moderate.

Do NOT reject an image merely because of:

- minor blur or low resolution;
- imperfect anatomy or object geometry;
- small artifacts;
- unrealistic details;
- mediocre aesthetics;
- imperfect composition, lighting, or colors;
- minor text rendering errors;
- moderate generation artifacts.

**Be permissive. When uncertain, KEEP the image.**

Return only the following JSON:
```json
{
  "decision": "KEEP" or "REJECT",
  "reason": "brief reason"
}
```
\end{Verbatim}
\end{tcolorbox}

The resulting \dialoguedataset{} contains 402,639 dialogues.
As shown in Fig.~\ref{fig:scenario_distributions}, Multi-Image Fusion, Image Editing, World Modeling, and Visual Reasoning account for 36\%, 25\%, 28\%, and 11\% of the training dialogues, respectively.
We retain this naturally varying training distribution while constructing a balanced benchmark with 100 dialogues from each scenario, enabling controlled comparison across scenarios.
Complete corpus- and scenario-level statistics are provided in Section~\ref{app:data_statistics}.

\subsection{Data Construction}
\label{app:dialogue_construction}
After data preparation, each scenario-specific pool is combined with scenario- and capability-oriented prompts to construct multi-turn multimodal stateful dialogues, as illustrated in Fig.~\ref{fig:pipeline}.b.
The complete data-construction prompts are provided in the supplementary material.
Within each dialogue, visual observations, textual instructions, and generated or edited images jointly determine how the multimodal dialogue state evolves.
The four scenarios instantiate distinct forms of this evolution.

\begin{enumerate}
    \item \textit{Multi-Image Fusion} : the assistant integrates multiple user-provided or previously generated images according to the dialogue history.
    \item \textit{Image Editing} : the assistant progressively edits an image while preserving prior modifications and visual consistency.
    \item \textit{World Modeling} : the assistant predicts and renders scene states under specified viewpoint changes or actions.
    \item \textit{Visual Reasoning} : the assistant generates visual outcomes for proposed hypotheses or solutions to support step-by-step reasoning.
\end{enumerate}

To make these interactions genuinely stateful, we incorporate six key state-management capabilities across the four scenarios: \textit{Coreference Resolution}, \textit{Dialogue-History Retrieval}, \textit{Cross-Turn Consistency}, \textit{Spatial and Physical Consistency}, \textit{Constraint Adherence}, and \textit{Correction-Based State Updating}.
\begin{enumerate}
    \item \textit{Coreference Resolution} grounds context-dependent expressions to the correct textual or visual entities.
    \item \textit{Dialogue-History Retrieval} identifies and uses information relevant to the current request from previous turns.
    \item \textit{Cross-Turn Consistency} preserves established visual attributes and prior edits across turns unless the user explicitly changes them.
    \item \textit{Spatial and Physical Consistency} maintains plausible geometry, physical relations, and state transitions.
    \item \textit{Constraint Adherence} consistently follows user-specified requirements and restrictions.
    \item \textit{Correction-Based State Updating} incorporates user corrections and replaces outdated information accordingly.
\end{enumerate}

These capabilities cut across the four scenarios rather than forming a fixed one-to-one mapping: multiple capabilities jointly construct and maintain the evolving state within a dialogue, and the same capability can occur in different scenarios.
Every dialogue in \dialoguedataset{} involves at least four of the six capabilities, requiring multiple cross-turn dependencies to be exercised jointly within the interaction.

\subsection{Automatic Verification}
\label{app:data_quality_control}
We apply quality control both before and after data construction, following the pipeline in Fig.~\ref{fig:pipeline}.
During data preparation, model-assisted filtering removes unsuitable visual materials and routes potentially repairable samples back through the preparation process.
After data construction, each dialogue is automatically verified for logical coherence and image--text consistency: the former examines the instructions and state transitions across turns, while the latter checks the visual content against its associated textual context.
Failed samples are iteratively corrected or regenerated and then re-verified, while dialogues that pass verification form \dialoguedataset{} or enter the benchmark-candidate pool.
This iterative process is intended to eliminate inconsistent state transitions, mismatched image--text pairs, and dialogues that do not realize the intended scenario or contextual dependencies.

\subsection{Human Verification and Benchmark Construction}
\label{app:benchmark_construction}
We construct \dialoguebench{} with LLM assistance from candidates that pass automatic verification, followed by exhaustive human verification and correction.
The benchmark is balanced across the four interaction scenarios, with 100 dialogues per scenario and 400 dialogues in total.
\textit{(i) Dialogue-state understanding QA pairs.} Each benchmark dialogue includes four QA pairs, one each for Reference Resolution, Context Retrieval, Constraint Tracking, and Correction Updating.
\textit{(ii) Fine-grained atomic criteria.} Each dialogue also includes atomic criteria for evaluating State Consistency in generated or edited visual outputs.
These atomic criteria decompose the requirements of a target image into independently verifiable checks, allowing State Consistency to be assessed separately from holistic Visual Quality.

As shown in the human-verification stage of Fig.~\ref{fig:pipeline}, the initial atomic checks are generated with LLM assistance and then reviewed together with the corresponding dialogue and visual content.
Three human annotators with graduate-level training in STEM disciplines verify each sample.
They inspect the dialogue logic, visual evidence, understanding questions, and atomic evaluation criteria, and correct samples that do not faithfully represent the intended state evolution or evaluation target.
Failed samples undergo manual correction or regeneration followed by another review, and every sample retained in the final benchmark is manually verified.
The final benchmark contains 1,600 understanding questions and 8,467 atomic visual checks, with detailed per-scenario statistics reported in Table~\ref{tab:appendix_benchmark_statistics}.

\begin{figure}[t]
\centering
\includegraphics[width=\textwidth]{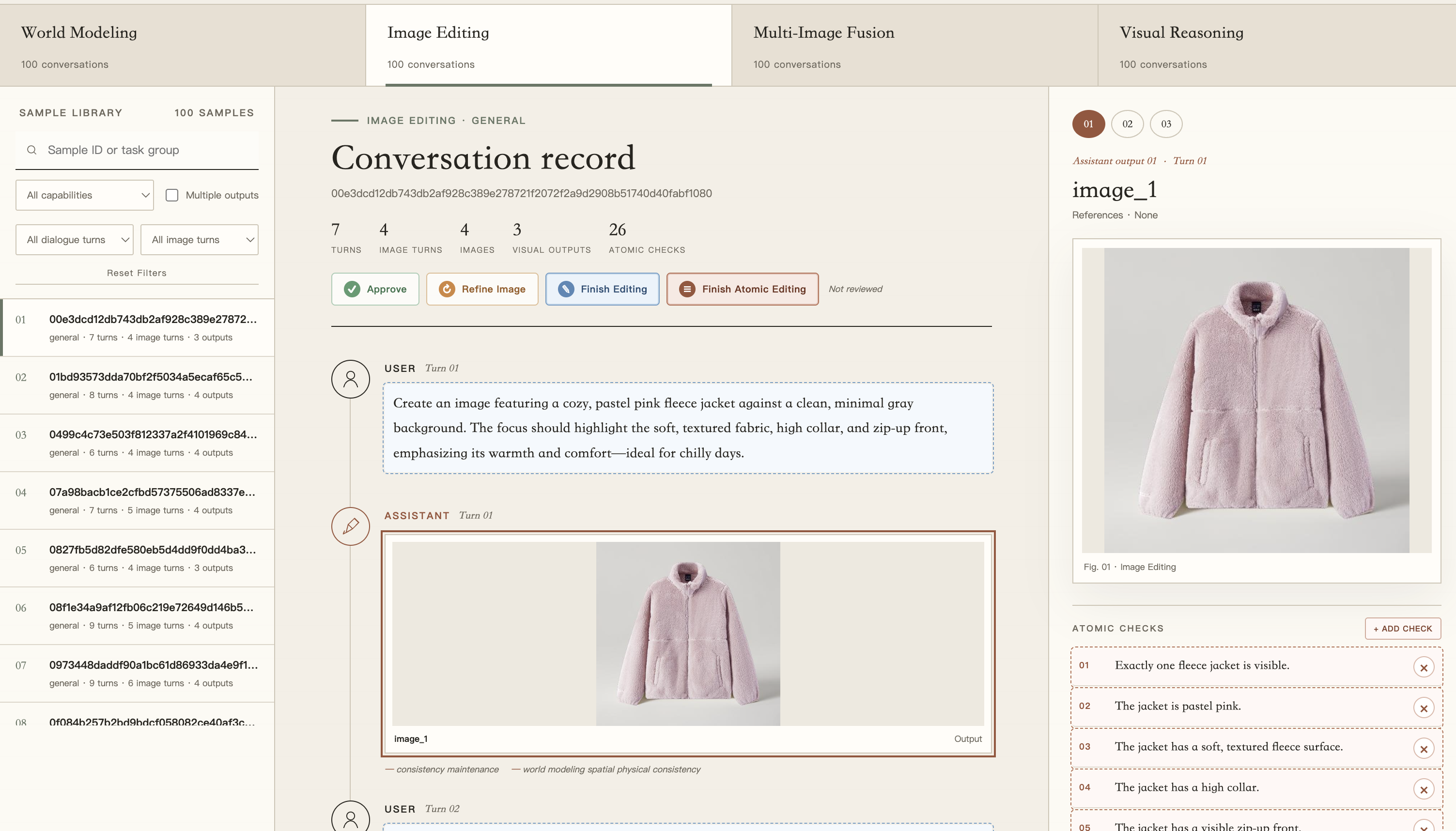}
\caption{\textbf{Human-verification interface for \dialoguebench{}.}
Annotators can browse the 100 benchmark dialogues in each interaction scenario, inspect the complete multi-turn context and visual outputs, and review or revise the corresponding atomic evaluation criteria.
The interface also records whether a sample is approved or requires image refinement, supporting the exhaustive verification and correction process used to construct the final benchmark.}
\label{fig:benchmark_verification_interface}
\end{figure}

Figure~\ref{fig:benchmark_verification_interface} shows the interface used to inspect complete dialogues, generated visual outputs, and their associated atomic checks during benchmark verification.

\subsection{Detailed Data Statistics}
\label{app:data_statistics}
\dialoguedataset{} contains 402,639 multi-turn dialogues, 2,047,236 images, and 2,643,291 user--assistant dialogue turns across the four interaction scenarios.
Among these dialogues, 321,868 contain at least four images, 215,027 contain at least five images, and 138,669 contain at least six images.
The dataset contains an average of 5.08 images and 6.56 dialogue turns per dialogue, with a maximum of 43 dialogue turns.
Table~\ref{tab:appendix_dataset_statistics} summarizes these statistics, while Fig.~\ref{fig:scenario_distributions} presents the distribution across the four interaction scenarios.

\dialoguebench{} contains 400 dialogues evenly distributed across the four scenarios, together with 1,883 images, 3,173 user--assistant dialogue turns, 1,600 dialogue-state understanding questions, and 8,467 atomic visual checks used in the benchmark evaluation.
Each dialogue contains one question for each of the four understanding capabilities, yielding 400 questions per capability.
Table~\ref{tab:appendix_benchmark_statistics} reports the corresponding scenario-level statistics.
\begin{table}[!htbp]
\centering
\caption{\textbf{Detailed statistics of \dialoguedataset{}.}
Each dialogue turn contains one user message and one assistant response; Total Messages therefore counts both sides of the interaction.
Scenario shares are rounded percentages corresponding to the training-set distribution shown in Fig.~\ref{fig:scenario_distributions}.
}
\label{tab:appendix_dataset_statistics}
\footnotesize
\setlength{\tabcolsep}{4pt}
\renewcommand{\arraystretch}{1.10}
\resizebox{\textwidth}{!}{%
\begin{tabular*}{\textwidth}{@{\extracolsep{\fill}}lrlrr@{}}
\toprule
\multicolumn{2}{c}{\textbf{Corpus and Sequence Statistics}} &
\multicolumn{3}{c}{\textbf{Dataset Distributions}} \\
\cmidrule(lr){1-2}\cmidrule(l){3-5}
\textbf{Statistic} & \textbf{Value} & \textbf{Images per Dialogue} & \textbf{Dialogues} & \textbf{Share} \\
\midrule
Total Dialogues & 402,639 & $<4$ & 80,771 & 20.06\% \\
Total Images & 2,047,236 & $=4$ & 106,841 & 26.54\% \\
Total Dialogue Turns & 2,643,291 & $=5$ & 76,358 & 18.96\% \\
Total User Messages & 2,643,291 & $\geq 4$ & 321,868 & 79.94\% \\
Total Assistant Messages & 2,643,291 & $\geq 5$ & 215,027 & 53.40\% \\
Total Messages & 5,286,582 & $\geq 6$ & 138,669 & 34.44\% \\
\addlinespace[2pt]
& & \multicolumn{3}{c}{\textit{Training Scenario Distribution}} \\
\cmidrule(l){3-5}
Average Images / Dialogue & 5.08 & Multi-Image Fusion & -- & 36\% \\
Average Turns / Dialogue & 6.56 & Image Editing & -- & 25\% \\
Average Messages / Dialogue & 13.13 & World Modeling & -- & 28\% \\
Maximum Turns / Dialogue & 43 & Visual Reasoning & -- & 11\% \\
Maximum Messages / Dialogue & 86 & Interaction Scenarios & 4 & -- \\
Images / Dialogue Turn & 0.77 & State-Management Capabilities & 6 & -- \\
Dialogues with $\geq 4$ Capabilities & 402,639 & Capability Coverage & -- & 100\% \\
\bottomrule
\end{tabular*}
}

\end{table}

\begin{table}[!htbp]
\centering
\caption{\textbf{Detailed statistics of \dialoguebench{} by interaction scenario.}
Images include all reference and target images contained in the benchmark dialogues.
Understanding QAs comprise Reference Resolution, Context Retrieval, Constraint Tracking, and Correction Updating questions, with one question of each type per dialogue.
Atomic Checks counts the independently verifiable visual criteria used in the benchmark evaluation and excludes capability-specific checks; Avg. Checks reports the average number of evaluated atomic checks per dialogue.}
\label{tab:appendix_benchmark_statistics}
\small
\setlength{\tabcolsep}{6pt}
\renewcommand{\arraystretch}{1.05}
\resizebox{\textwidth}{!}{%
\begin{tabular}{@{}lrrrrrr@{}}
\toprule
\textbf{Scenario} & \textbf{Dialogues} & \textbf{Images} & \textbf{Dialogue Turns} & \textbf{Understanding QAs} & \textbf{Atomic Checks} & \textbf{Avg. Checks} \\
\midrule
Multi-Image Fusion & 100 & 400 & 777 & 400 & 1,890 & 18.90 \\
Image Editing & 100 & 499 & 754 & 400 & 3,408 & 34.08 \\
World Modeling & 100 & 467 & 598 & 400 & 1,141 & 11.41 \\
Visual Reasoning & 100 & 517 & 1,044 & 400 & 2,028 & 20.28 \\
\midrule
\textbf{Total} & \textbf{400} & \textbf{1,883} & \textbf{3,173} & \textbf{1,600} & \textbf{8,467} & \textbf{21.17} \\
\bottomrule
\end{tabular}%
}

\end{table}

\section{Evaluation Protocol and Metrics}
\label{app:evaluation}

\subsection{Dialogue-State Understanding Evaluation}
\label{app:understanding_evaluation}
Each benchmark dialogue contains four diagnostic questions, one for each explicitly testable understanding capability: Reference Resolution, Context Retrieval, Constraint Tracking, and Correction Updating.
The evaluator receives the dialogue context, the diagnostic question, a reference answer, and the model prediction, and judges semantic correctness rather than exact lexical overlap.
A prediction receives credit only when it contains all information required by the reference answer without a material factual error or contradiction.
Equivalent wording and correct additional detail are accepted, whereas partially correct, unsupported, or materially incomplete answers are marked incorrect.
The score for each capability is the proportion of correctly answered questions, and the overall dialogue-state understanding score is the arithmetic mean defined in Appendix~\ref{app:score_calculation}.

\subsection{Visual Quality and State Consistency Evaluation}
\label{app:visual_evaluation}
State Consistency decomposes the expected target state into independently verifiable atomic visual checks.
For each generated image, the evaluator compares the candidate with the ground-truth target and assigns a binary outcome to every atomic check.
A check receives credit only when all of its required conditions are visibly satisfied; semantically equivalent visual realizations are accepted when they do not alter the evaluated state property.
The State Consistency score is the fraction of atomic checks that receive credit.
Capability-specific checks are evaluated separately and are used only for analyses that require alignment between an understanding item and its corresponding generated outcome.

Visual Quality assesses the generated image independently of task correctness on a 0--10 integer scale.
The rubric considers structural coherence, visual fidelity, generation artifacts, and overall naturalness, with higher scores indicating better perceptual quality.
The complete prompts for all three evaluators are reproduced in Appendix~\ref{app:judge_template}.

\subsection{Score Calculation}
\label{app:score_calculation}
For each interaction scenario, we first compute the dialogue-state understanding score as the arithmetic mean of the four understanding metrics: Reference Resolution (RR), Context Retrieval (CR), Constraint Tracking (CT), and Correction Updating (CU).
Specifically, the understanding score is calculated as
\begin{equation}
S_{\mathrm{understanding}}
=\frac{S_{\mathrm{RR}}+S_{\mathrm{CR}}+S_{\mathrm{CT}}+S_{\mathrm{CU}}}{4}.
\end{equation}
Before aggregation, all component scores are converted to a 0--100 scale: the State Consistency (SC) score, originally expressed as a fraction, is multiplied by 100, and the Visual Quality (VQ) score, originally measured on a 0--10 scale, is multiplied by 10.
For models with both understanding and generation capabilities, the final score for each scenario is computed using a $4{:}4{:}2$ weighting over dialogue-state understanding, State Consistency, and Visual Quality:
\begin{equation}
S_{\mathrm{scene}}
=0.4S_{\mathrm{understanding}}
+0.4S_{\mathrm{SC}}
+0.2S_{\mathrm{VQ}}.
\end{equation}
For models without dialogue-state understanding capabilities, the scenario score is computed from State Consistency and Visual Quality using a $6{:}4$ weighting:
\begin{equation}
S_{\mathrm{scene}}
=0.6S_{\mathrm{SC}}
+0.4S_{\mathrm{VQ}}.
\end{equation}
This weighting ensures that state faithfulness remains the primary component of the final score while retaining an explicit assessment of perceptual visual quality.

\subsection{History and Input Formatting}
\label{app:history_input_format}
Our main evaluation adopts the GT-History setting to measure the model's ability to use accumulated dialogue state without conflating it with errors produced in earlier generation turns.
When evaluating a target generation or editing turn, we retain the complete textual dialogue preceding that turn and replace all earlier assistant-generated images with their corresponding ground-truth images.
User-provided reference images remain unchanged.
The target output itself is never included in the input context.

We additionally evaluate Self-History to approximate realistic multi-turn deployment.
Under this setting, each image produced by the evaluated model is inserted into the dialogue history and used as visual context for subsequent turns, while the textual history remains unchanged.
User-provided reference images again remain unchanged.
Consequently, errors in earlier visual outputs may alter the context available at later turns and accumulate throughout the interaction.

We use three input organizations according to the interfaces supported by individual models.
For interleaved image--text input, denoted as \textit{yes-first}, each image is inserted at the position of its first mention in the dialogue, preserving the original temporal order of the interaction.
For image-first input, denoted as \textit{yes-front}, all visual inputs available at the target turn are placed before the textual dialogue in their order of appearance.
For models that cannot process sequential multi-image inputs, the available images are concatenated in dialogue order into a single composite visual input and supplied together with the corresponding textual instruction.
The input organization used for each model is reported in Table~\ref{tab:main_results}.

\subsection{Baseline Inference Configuration}
\label{app:baseline_inference}
All models are evaluated on the same set of dialogues and target generation or editing turns in \dialoguebench{}.
For each target turn, the input contains the preceding multi-turn dialogue and all reference images available under the selected history setting described above.
Models with an understanding interface are additionally evaluated on the 1,600 dialogue-state understanding questions, while generation- or editing-only models are marked as not applicable for this component.
We request an aspect-ratio-preserving output frame whose long edge is 512 pixels, with both dimensions rounded to multiples of 16; when a native model interface imposes a fixed or internally selected resolution, we retain its native output for evaluation.
We use a global seed of 42 and derive a deterministic seed for each sample from its identifier.
Text responses are decoded greedily with at most 512 new tokens, unless the model's official interface requires a different decoding procedure.
For closed-source systems, implementation-specific decoding parameters are unavailable, so we use their provided interfaces with the same target-turn context and output requirements.
Dialogue-state understanding, State Consistency, and Visual Quality are evaluated with Qwen-3.8-Flash-Next using deterministic decoding ($\text{temperature}=0$) with explicit reasoning disabled and the three prompts reproduced in Appendix~\ref{app:judge_template}.
Table~\ref{tab:baseline_inference} summarizes the model-specific input handling and the principal generation parameters used in our evaluation.

\begin{table}[t]
\centering
\caption{\textbf{Baseline inference configurations used on \dialoguebench.}
Only settings that materially affect input organization or output generation are reported; all models otherwise follow the shared evaluation protocol described in the text.}
\label{tab:baseline_inference}
\scriptsize
\setlength{\tabcolsep}{3.2pt}
\renewcommand{\arraystretch}{1.10}
\begin{tabular}{@{}>{\raggedright\arraybackslash}p{0.15\textwidth}
                    >{\raggedright\arraybackslash}p{0.21\textwidth}
                    >{\raggedright\arraybackslash}p{0.25\textwidth}
                    >{\raggedright\arraybackslash}p{0.31\textwidth}@{}}
\toprule
\textbf{Model} & \textbf{Generation mechanism} & \textbf{Visual-input handling} & \textbf{Principal inference settings} \\
\midrule
BAGEL~\citep{deng2025emerging} and +\dialoguedataset{}
& Unified MoT for understanding and generation
& Native interleaved image--text sequence; no image concatenation
& 50 steps; text/image CFG $4.0/1.5$; parallel CFG; timestep shift $3.0$ \\

Emu3.5~\citep{cui2025emu3}
& Unified autoregressive generation with discrete visual tokens
& Native interleaved image--text sequence; no image concatenation
& Image/text guidance $3.0/1.0$; image top-$k$ 10,240; top-$p$ $1.0$ \\

OmniGen2~\citep{wu2025omnigen2}
& Unified diffusion pipeline
& Separate visual inputs; references are concatenated into a grid only when more than three are present
& 50 steps; text guidance $4.0$; resolution alignment disabled \\

Ovis-U1~\citep{wang2025ovis}
& Unified MLLM with a diffusion generation head
& Ordered image placeholders; multiple references are concatenated into a grid
& 50 steps; text/image CFG $5.0/1.5$ \\

InternVL-U~\citep{tian2026internvl}
& Unified VLM with diffusion-based generation
& Multiple references are concatenated into a grid
& 50 steps; deterministic per-sample seed \\

Janus-Pro~\citep{chen2025janus}
& Autoregressive generation with discrete visual tokens
& Interleaved inputs for understanding; generation uses the flattened dialogue text without reference-image conditioning
& 576 visual tokens per image; CFG $5.0$; greedy decoding \\

Show-o2~\citep{xie2026show}
& Continuous-token flow-matching generation
& Reference-image latents are concatenated before the target latent
& 50 Euler steps; guidance $7.5$; time-shifting factor $3.0$ \\

UniAR~\citep{peng2026unified}
& Unified autoregressive model with a visual decoder
& Native interleaving for understanding; multiple generation references are concatenated into a grid
& CFG $1.5$; greedy text decoding \\

OmniGen~\citep{xiao2025omnigen}
& Diffusion-based image generation and editing
& Ordered image placeholders; no explicit reference-image limit
& Text/image guidance $2.5/1.6$ \\

FLUX.2 [dev]~\citep{labs2025flux2}
& Rectified-flow diffusion transformer
& Up to ten separate references; larger sets are concatenated into a grid
& 50 steps; guidance $4.0$ \\

Qwen-Edit~\citep{wu2025qwen}
& Multi-image-conditioned diffusion transformer
& Native multi-image conditioning without concatenation
& 40 steps; true CFG $4.0$; guidance $1.0$ \\

Step1X-v1p2~\citep{liu2025step1x}
& Diffusion-transformer image editing
& Multiple references are concatenated into a grid
& 50 steps; true CFG $6.0$; reflection and thinking modes disabled \\

UniPic 3.0~\citep{wei2026skywork}
& Multi-image composition and editing
& Multiple references are concatenated into a grid
& 50 steps; true CFG $4.0$; guidance $1.0$ \\
\bottomrule
\end{tabular}

\end{table}

\subsection{Understanding--Realization Gap Evaluation}
\label{app:gap_evaluation}
We measure the understanding--realization gap by conditioning visual-realization evaluation on correct dialogue-state understanding.
For each dialogue and evaluated state property, such as a persistent user constraint or a correction-based state update, we first identify the corresponding understanding question and its aligned generation-side capability checks.
To ensure a fair model comparison, we use a shared evaluation subset rather than selecting correctly understood items separately for each model.
Specifically, a dialogue is retained for capability $c$ only if every evaluated model receives full credit on its corresponding understanding question and has at least one aligned generation-side check.
This prevents a model with weaker understanding from being evaluated on an easier or smaller model-specific subset and ensures that all reported gaps are computed on exactly the same dialogue--capability pairs.

Let $\mathcal{M}$ denote the set of evaluated models, let $u_{m,d,c}\in[0,1]$ be the understanding score of model $m$ for capability $c$ in dialogue $d$, and let $\mathcal{A}_{m,d,c}$ be its aligned generation-side checks.
For each capability, the shared eligible dialogue set is
\begin{equation}
\mathcal{E}_{c}
=\left\{d\,\middle|\,
u_{m,d,c}=1\ \land\ |\mathcal{A}_{m,d,c}|>0,
\ \forall m\in\mathcal{M}\right\}.
\end{equation}
Each aligned check $a\in\mathcal{A}_{m,d,c}$ receives a binary score $y_{m,d,c,a}$, where a correctly realized state property receives 1 and an incorrectly realized property receives 0.
For example, when the model receives full credit for understanding an active user constraint, each corresponding constraint-keeping check in the generated image contributes 1 if satisfied and 0 otherwise.
The visual-realization score for a retained dialogue--capability pair is
\begin{equation}
R_{m,d,c}
=\frac{1}{|\mathcal{A}_{m,d,c}|}
\sum_{a\in\mathcal{A}_{m,d,c}} y_{m,d,c,a}.
\end{equation}
Because the atomic scores are binary, $R_{m,d,c}$ is naturally normalized to $[0,1]$.
We first average within each dialogue--capability pair and then across the shared eligible dialogues so that dialogues containing more checks do not receive disproportionate weight:
\begin{equation}
R_{m,c}
=\frac{1}{|\mathcal{E}_{c}|}
\sum_{d\in\mathcal{E}_{c}} R_{m,d,c}.
\end{equation}
We apply this procedure to the aligned Constraint Tracking and Correction Updating understanding items and their corresponding Constraint Keeping and Correction-Based State Updating checks on the generation side.
Across the nine UMMs included in this analysis, all models correctly answer 78 Constraint Tracking items and 71 Correction Updating items; after requiring aligned generation checks for every model, the shared subsets contain 77 and 66 dialogues, respectively.
The capability-specific understanding--realization gap is defined as
\begin{equation}
G_{m,c}=1-R_{m,c},
\end{equation}
where a larger value indicates a greater failure to visually realize state properties that the model has correctly understood.
The overall gap is the macro-average of the two capability-specific gaps.

\section{Training Details}
\label{app:training_details}
We fine-tune BAGEL-7B-MoT on \dialoguedataset{} using NVIDIA A800 and A100 GPUs, each with 80 GB of memory.
Our compute resources comprise a six-GPU A800 configuration and a 16-GPU A100 configuration; the provided launch configuration uses six processes on a single node.
We employ fully sharded data parallelism (FSDP) under the \texttt{FULL\_SHARD} strategy.
The model uses \texttt{Qwen2MoTDecoderLayer} and is initialized from the Hugging Face checkpoint and its exponential moving average weights, with only model parameters restored.
We jointly optimize visual generation and visual understanding objectives.
The VAE is frozen, whereas the language model, vision encoder, and understanding module remain trainable.
We use a constant learning rate of $1\times10^{-5}$ with 100 warm-up steps, a cross-entropy loss weight of 0.25, an MSE loss weight of 1.0, gradient accumulation of 1, and a maximum gradient norm of 1.0.
The expected token count is 12,288, while both the overall and per-sample maximum token counts are 16,384.
The maximum latent size is 64, and FlexAttention is enabled.
Data loading uses one worker per process and a prefetch factor of 2.

\begin{table}[t]
\centering
\caption{\textbf{Training configuration.} Settings used to fine-tune BAGEL-7B-MoT on \dialoguedataset, including the model and system setup, optimization hyperparameters, and sequence-length settings.}
\label{tab:training_details}
\footnotesize
\setlength{\tabcolsep}{5pt}
\renewcommand{\arraystretch}{1.14}
\begin{tabular}{@{}>{\raggedright\arraybackslash}p{0.17\textwidth}
                    >{\raggedright\arraybackslash}p{0.27\textwidth}
                    >{\raggedright\arraybackslash}p{0.17\textwidth}
                    >{\raggedright\arraybackslash}p{0.27\textwidth}@{}}
\toprule
\multicolumn{2}{c}{\textbf{Model and System Configuration}} &
\multicolumn{2}{c}{\textbf{Optimization and Sequence Configuration}} \\
\cmidrule(lr){1-2}\cmidrule(l){3-4}
\textbf{Setting} & \textbf{Value} & \textbf{Setting} & \textbf{Value} \\
\midrule
Base model & BAGEL-7B-MoT & Learning rate & $1\times10^{-5}$ \\
Decoder layer & \texttt{Qwen2MoTDecoderLayer} & Scheduler & Constant \\
Compute resources & $6\times$ NVIDIA A800 80GB; $16\times$ NVIDIA A100 80GB & Warm-up steps & 100 \\
Launch configuration & 1 node, 6 processes & CE / MSE loss weights & 0.25 / 1.0 \\
Sharding strategy & FSDP \texttt{FULL\_SHARD} & Gradient accumulation steps & 1 \\
Trainable modules & LLM, ViT, understanding module & Maximum gradient norm & 1.0 \\
Frozen module & VAE & Expected token count & 12,288 \\
Training objectives & Visual understanding and generation & Maximum token count & 16,384 \\
Attention implementation & FlexAttention & Maximum tokens per sample & 16,384 \\
Data-loader configuration & 1 worker per process; prefetch factor 2 & Maximum latent size & 64 \\
\bottomrule
\end{tabular}

\end{table}

\section{Complete Experimental Results}
\label{app:complete_experimental_results}

\subsection{Aggregate Results on \dialoguebench{}}
\label{app:complete_benchmark_results}
Complete aggregate results for all evaluated models and interaction scenarios are reported in Table~\ref{tab:main_results}.
Scores are computed from the available valid evaluation records; missing records are not assigned zero scores.
For GPT-Image-2, scores are computed over 948 valid outputs; four requests rejected by the upstream content policy are excluded.
For Nano Banana 2, scores are computed over 939 valid outputs.

\subsection{Dialogue-State Understanding Results}
\label{app:full_understanding_results}
For each interaction scenario, we report two complementary types of tables.
The first reports the four capability-level dialogue-state understanding scores and their arithmetic mean.
The second reports the average understanding score alongside the generation-side State Consistency and Visual Quality scores, followed by the weighted overall average.
Models that only support image generation or editing do not produce answers for the understanding evaluation and are therefore marked with dashes in the corresponding columns.
Table~\ref{tab:overall_understanding_by_scenario} summarizes dialogue-state understanding performance across the four interaction scenarios.

\begin{table}[t]
\centering
\caption{\textbf{Dialogue-state understanding results across the four interaction scenarios.}
Fusion, Editing, Modeling, and Reasoning denote Multi-Image Fusion, Image Editing, World Modeling, and Visual Reasoning, respectively.
Each scenario score is the arithmetic mean of RR, CR, CT, and CU, and Avg. is the macro-average across the four scenarios.
All results are reported on a 0--100 scale, and higher values are better.}
\label{tab:overall_understanding_by_scenario}
\small
\setlength{\tabcolsep}{9pt}
\renewcommand{\arraystretch}{1.0}
\begin{tabular}{@{}lccccc@{}}
\toprule
\textbf{Model} & \textbf{Fusion} & \textbf{Editing} & \textbf{Modeling} & \textbf{Reasoning} & \textbf{Avg.} \\
\midrule
OmniGen2~\citep{wu2025omnigen2} & 46.75 & 50.00 & 65.00 & 42.75 & 51.13 \\
Janus-Pro~\citep{chen2025janus} & 68.00 & 41.75 & 69.25 & 51.50 & 57.63 \\
Show-o2~\citep{xie2026show} & 74.00 & 64.25 & 79.50 & 54.00 & 67.94 \\
Ovis-U1~\citep{wang2025ovis} & 83.75 & 25.25 & 65.75 & 55.25 & 57.50 \\
Emu3.5~\citep{cui2025emu3} & 80.00 & 47.50 & 85.50 & 71.00 & 71.00 \\
InternVL-U~\citep{tian2026internvl} & 82.75 & 73.00 & 88.75 & 63.00 & 76.88 \\
UniAR~\citep{peng2026unified} & 75.50 & 51.50 & 73.75 & 60.25 & 65.25 \\
\midrule
BAGEL~\citep{deng2025emerging} & 85.75 & 76.50 & 85.25 & 78.25 & 81.44 \\
\textbf{+ \dialoguedataset{}} & \textbf{96.25} & \textbf{98.00} & \textbf{95.50} & \textbf{93.00} & \textbf{95.69} \\
\bottomrule
\end{tabular}

\end{table}

\begin{table}[!htbp]
\centering
\caption{\textbf{Dialogue-state understanding results for Visual Reasoning.}
RR, CR, CT, and CU denote Reference Resolution, Context Retrieval, Constraint Tracking, and Correction Updating, respectively.
DSU Avg. is the arithmetic mean of the four capability scores.
All results are reported on a 0--100 scale, and higher values are better.
An upward arrow indicates an improvement over BAGEL.
}
\label{tab:visual_reasoning_understanding}
\small
\setlength{\tabcolsep}{7pt}
\renewcommand{\arraystretch}{0.98}
\resizebox{\textwidth}{!}{%
\begin{tabular}{@{}clcccccccc@{}}
\toprule
\textbf{Index} & \textbf{Model} & \textbf{Size} & \textbf{Modality} & \textbf{Format} & \textbf{RR} & \textbf{CR} & \textbf{CT} & \textbf{CU} & \textbf{DSU Avg.} \\
\midrule
1 & OmniGen2~\citep{wu2025omnigen2} & 7B & \tableicon{gen_and_com} & \tableicon{image_then_text} & 15.0 & 66.0 & 28.0 & 62.0 & 42.75 \\
2 & Janus-Pro~\citep{chen2025janus} & 7B & \tableicon{gen_and_com} & \tableicon{image_then_text} & 30.0 & 71.0 & 41.0 & 64.0 & 51.50 \\
3 & Show-o2~\citep{xie2026show} & 7B & \tableicon{gen_and_com} & \tableicon{concat} & 18.0 & 81.0 & 43.0 & 74.0 & 54.00 \\
4 & Ovis-U1~\citep{wang2025ovis} & 3B & \tableicon{gen_and_com} & \tableicon{concat} & 34.0 & 70.0 & 42.0 & 75.0 & 55.25 \\
5 & Emu3.5~\citep{cui2025emu3} & 34B & \tableicon{gen_and_com} & \tableicon{image_then_text} & 34.0 & 87.0 & 71.0 & 92.0 & 71.00 \\
6 & InternVL-U~\citep{tian2026internvl} & 4B & \tableicon{gen_and_com} & \tableicon{concat} & 25.0 & 90.0 & 53.0 & 84.0 & 63.00 \\
7 & UniAR~\citep{peng2026unified} & 10B & \tableicon{gen_and_com} & \tableicon{image_then_text} & 41.0 & 84.0 & 37.0 & 79.0 & 60.25 \\
8 & BAGEL~\citep{deng2025emerging} & 7B & \tableicon{gen_and_com} & \tableicon{image_then_text} & 59.0 & 95.0 & 65.0 & 94.0 & 78.25 \\
9 & \textbf{+ \dialoguedataset{}} & 7B & \tableicon{gen_and_com} & \tableicon{image_then_text} & \textbf{86.0}\,$\uparrow$ & \textbf{98.0}\,$\uparrow$ & \textbf{90.0}\,$\uparrow$ & \textbf{98.0}\,$\uparrow$ & \textbf{93.00}\,$\uparrow$ \\
\bottomrule
\end{tabular}%
}

\end{table}

\begin{table}[!htbp]
\centering
\caption{\textbf{Understanding and visual-realization results for Visual Reasoning.}
SC and VQ denote State Consistency and Visual Quality, respectively.
All component and aggregate results are reported on a 0--100 scale; VQ scores are converted from their original 0--10 scale by multiplying by 10.
Overall Avg. is computed using the weighting defined in Appendix~\ref{app:score_calculation}.
Image-generation-only models do not provide DSU scores; a dash denotes an unsupported or unavailable component.
Modality and input-format icons follow the conventions in Table~\ref{tab:main_results}.
An upward arrow indicates an improvement over BAGEL.
}
\label{tab:visual_reasoning_detailed}
\small
\setlength{\tabcolsep}{7pt}
\renewcommand{\arraystretch}{0.98}
\resizebox{\textwidth}{!}{%
\begin{tabular}{@{}clccccccc@{}}
\toprule
\textbf{Index} & \textbf{Model} & \textbf{Size} & \textbf{Modality} & \textbf{Format} & \textbf{DSU Avg.} & \textbf{SC} & \textbf{VQ} & \textbf{Overall Avg.} \\
\midrule
1 & GPT-Image-2~\citep{openai2026gptimage2} & -- & \tableicon{image_only} & \tableicon{image_first} & -- & 75.2 & 70.8 & 73.46 \\
2 & Nano Banana 2~\citep{google2026nanobanana2} & -- & \tableicon{image_only} & \tableicon{image_then_text} & -- & 80.7 & 71.3 & 76.93 \\
\midrule
3 & Step1X-v1p2~\citep{liu2025step1x} & 12B & \tableicon{image_only} & \tableicon{concat} & -- & 54.0 & 39.0 & 48.00 \\
4 & Qwen-Edit~\citep{wu2025qwen} & 20B & \tableicon{image_only} & \tableicon{image_then_text} & -- & 39.4 & 35.2 & 37.72 \\
5 & FLUX.2 [dev]~\citep{labs2025flux2} & 32B & \tableicon{image_only} & \tableicon{concat} & -- & 49.0 & \textbf{68.2} & 56.68 \\
6 & UniPic 3.0~\citep{wei2026skywork} & 20B & \tableicon{image_only} & \tableicon{concat} & -- & 56.1 & 37.9 & 48.82 \\
7 & OmniGen~\citep{xiao2025omnigen} & 3.8B & \tableicon{image_only} & \tableicon{image_then_text} & -- & 28.1 & 42.7 & 33.94 \\
\midrule
8 & OmniGen2~\citep{wu2025omnigen2} & 7B & \tableicon{gen_and_com} & \tableicon{image_then_text} & 42.75 & 54.8 & 42.9 & 47.62 \\
9 & Janus-Pro~\citep{chen2025janus} & 7B & \tableicon{gen_and_com} & \tableicon{image_then_text} & 51.50 & 7.4 & 41.5 & 31.86 \\
10 & Show-o2~\citep{xie2026show} & 7B & \tableicon{gen_and_com} & \tableicon{concat} & 54.00 & 8.5 & 22.1 & 29.42 \\
11 & Ovis-U1~\citep{wang2025ovis} & 3B & \tableicon{gen_and_com} & \tableicon{concat} & 55.25 & 32.9 & 40.9 & 43.42 \\
12 & Emu3.5~\citep{cui2025emu3} & 34B & \tableicon{gen_and_com} & \tableicon{image_then_text} & 71.00 & 50.4 & 62.1 & 60.98 \\
13 & InternVL-U~\citep{tian2026internvl} & 4B & \tableicon{gen_and_com} & \tableicon{concat} & 63.00 & 34.8 & 37.0 & 46.52 \\
14 & UniAR~\citep{peng2026unified} & 10B & \tableicon{gen_and_com} & \tableicon{image_then_text} & 60.25 & 16.0 & 27.2 & 35.92 \\
15 & BAGEL~\citep{deng2025emerging} & 7B & \tableicon{gen_and_com} & \tableicon{image_then_text} & 78.25 & 47.0 & 59.9 & 62.06 \\
16 & \textbf{+ \dialoguedataset{}} & 7B & \tableicon{gen_and_com} & \tableicon{image_then_text} & \textbf{93.00}\,$\uparrow$ & \textbf{64.1}\,$\uparrow$ & 64.8\,$\uparrow$ & \textbf{75.80}\,$\uparrow$ \\
\bottomrule
\end{tabular}%
}

\end{table}

\begin{table}[!htbp]
\centering
\caption{\textbf{Dialogue-state understanding results for Image Editing.}
RR, CR, CT, and CU denote Reference Resolution, Context Retrieval, Constraint Tracking, and Correction Updating, respectively.
DSU Avg. is the arithmetic mean of the four capability scores.
All results are reported on a 0--100 scale, and higher values are better.
An upward arrow indicates an improvement over BAGEL.
}
\label{tab:image_editing_understanding}
\small
\setlength{\tabcolsep}{7pt}
\renewcommand{\arraystretch}{0.98}
\resizebox{\textwidth}{!}{%
\begin{tabular}{@{}clcccccccc@{}}
\toprule
\textbf{Index} & \textbf{Model} & \textbf{Size} & \textbf{Modality} & \textbf{Format} & \textbf{RR} & \textbf{CR} & \textbf{CT} & \textbf{CU} & \textbf{DSU Avg.} \\
\midrule
1 & OmniGen2~\citep{wu2025omnigen2} & 7B & \tableicon{gen_and_com} & \tableicon{image_then_text} & 73.0 & 44.0 & 44.0 & 39.0 & 50.00 \\
2 & Janus-Pro~\citep{chen2025janus} & 7B & \tableicon{gen_and_com} & \tableicon{image_then_text} & 18.0 & 55.0 & 62.0 & 32.0 & 41.75 \\
3 & Show-o2~\citep{xie2026show} & 7B & \tableicon{gen_and_com} & \tableicon{concat} & 70.0 & 60.0 & 75.0 & 52.0 & 64.25 \\
4 & Ovis-U1~\citep{wang2025ovis} & 3B & \tableicon{gen_and_com} & \tableicon{concat} & 32.0 & 25.0 & 35.0 & 9.0 & 25.25 \\
5 & Emu3.5~\citep{cui2025emu3} & 34B & \tableicon{gen_and_com} & \tableicon{image_then_text} & 2.0 & 66.0 & 81.0 & 41.0 & 47.50 \\
6 & InternVL-U~\citep{tian2026internvl} & 4B & \tableicon{gen_and_com} & \tableicon{concat} & 67.0 & 83.0 & 87.0 & 55.0 & 73.00 \\
7 & UniAR~\citep{peng2026unified} & 10B & \tableicon{gen_and_com} & \tableicon{image_then_text} & 66.0 & 63.0 & 47.0 & 30.0 & 51.50 \\
8 & BAGEL~\citep{deng2025emerging} & 7B & \tableicon{gen_and_com} & \tableicon{image_then_text} & 85.0 & 80.0 & 82.0 & 59.0 & 76.50 \\
9 & \textbf{+ \dialoguedataset{}} & 7B & \tableicon{gen_and_com} & \tableicon{image_then_text} & \textbf{98.0}\,$\uparrow$ & \textbf{97.0}\,$\uparrow$ & \textbf{98.0}\,$\uparrow$ & \textbf{99.0}\,$\uparrow$ & \textbf{98.00}\,$\uparrow$ \\
\bottomrule
\end{tabular}%
}

\end{table}

\begin{table}[!htbp]
\centering
\caption{\textbf{Understanding and visual-realization results for Image Editing.}
SC and VQ denote State Consistency and Visual Quality, respectively.
All component and aggregate results are reported on a 0--100 scale; VQ scores are converted from their original 0--10 scale by multiplying by 10.
Overall Avg. is computed using the weighting defined in Appendix~\ref{app:score_calculation}.
Image-generation-only models do not provide DSU scores; a dash denotes an unsupported or unavailable component.
Modality and input-format icons follow the conventions in Table~\ref{tab:main_results}.
An upward arrow indicates an improvement over BAGEL.
}
\label{tab:image_editing_detailed}
\small
\setlength{\tabcolsep}{7pt}
\renewcommand{\arraystretch}{0.98}
\resizebox{\textwidth}{!}{%
\begin{tabular}{@{}clccccccc@{}}
\toprule
\textbf{Index} & \textbf{Model} & \textbf{Size} & \textbf{Modality} & \textbf{Format} & \textbf{DSU Avg.} & \textbf{SC} & \textbf{VQ} & \textbf{Overall Avg.} \\
\midrule
1 & GPT-Image-2~\citep{openai2026gptimage2} & -- & \tableicon{image_only} & \tableicon{image_first} & -- & 93.6 & 85.9 & 90.54 \\
2 & Nano Banana 2~\citep{google2026nanobanana2} & -- & \tableicon{image_only} & \tableicon{image_then_text} & -- & 91.1 & 81.6 & 87.32 \\
\midrule
3 & Step1X-v1p2~\citep{liu2025step1x} & 12B & \tableicon{image_only} & \tableicon{concat} & -- & 64.0 & 62.4 & 63.36 \\
4 & Qwen-Edit~\citep{wu2025qwen} & 20B & \tableicon{image_only} & \tableicon{image_then_text} & -- & 53.7 & 56.2 & 54.70 \\
5 & FLUX.2 [dev]~\citep{labs2025flux2} & 32B & \tableicon{image_only} & \tableicon{concat} & -- & 73.3 & \textbf{76.4} & 74.54 \\
6 & UniPic 3.0~\citep{wei2026skywork} & 20B & \tableicon{image_only} & \tableicon{concat} & -- & 68.3 & 74.2 & 70.66 \\
7 & OmniGen~\citep{xiao2025omnigen} & 3.8B & \tableicon{image_only} & \tableicon{image_then_text} & -- & 55.4 & 62.2 & 58.12 \\
\midrule
8 & OmniGen2~\citep{wu2025omnigen2} & 7B & \tableicon{gen_and_com} & \tableicon{image_then_text} & 50.00 & 51.6 & 55.3 & 51.70 \\
9 & Janus-Pro~\citep{chen2025janus} & 7B & \tableicon{gen_and_com} & \tableicon{image_then_text} & 41.75 & 31.7 & 62.7 & 41.94 \\
10 & Show-o2~\citep{xie2026show} & 7B & \tableicon{gen_and_com} & \tableicon{concat} & 64.25 & 30.8 & 35.6 & 45.12 \\
11 & Ovis-U1~\citep{wang2025ovis} & 3B & \tableicon{gen_and_com} & \tableicon{concat} & 25.25 & 59.4 & 62.0 & 46.24 \\
12 & Emu3.5~\citep{cui2025emu3} & 34B & \tableicon{gen_and_com} & \tableicon{image_then_text} & 47.50 & 67.0 & 71.5 & 60.10 \\
13 & InternVL-U~\citep{tian2026internvl} & 4B & \tableicon{gen_and_com} & \tableicon{concat} & 73.00 & 59.2 & 66.2 & 66.12 \\
14 & UniAR~\citep{peng2026unified} & 10B & \tableicon{gen_and_com} & \tableicon{image_then_text} & 51.50 & 32.7 & 40.7 & 41.82 \\
15 & BAGEL~\citep{deng2025emerging} & 7B & \tableicon{gen_and_com} & \tableicon{image_then_text} & 76.50 & 61.5 & 60.9 & 67.40 \\
16 & \textbf{+ \dialoguedataset{}} & 7B & \tableicon{gen_and_com} & \tableicon{image_then_text} & \textbf{98.00}\,$\uparrow$ & 79.1\,$\uparrow$ & 70.6\,$\uparrow$ & \textbf{84.96}\,$\uparrow$ \\
\bottomrule
\end{tabular}%
}

\end{table}

\begin{table}[!htbp]
\centering
\caption{\textbf{Dialogue-state understanding results for World Modeling.}
RR, CR, CT, and CU denote Reference Resolution, Context Retrieval, Constraint Tracking, and Correction Updating, respectively.
DSU Avg. is the arithmetic mean of the four capability scores.
All results are reported on a 0--100 scale, and higher values are better.
An upward arrow indicates an improvement over BAGEL.
}
\label{tab:world_modeling_understanding}
\small
\setlength{\tabcolsep}{7pt}
\renewcommand{\arraystretch}{0.98}
\resizebox{\textwidth}{!}{%
\begin{tabular}{@{}clcccccccc@{}}
\toprule
\textbf{Index} & \textbf{Model} & \textbf{Size} & \textbf{Modality} & \textbf{Format} & \textbf{RR} & \textbf{CR} & \textbf{CT} & \textbf{CU} & \textbf{DSU Avg.} \\
\midrule
1 & OmniGen2~\citep{wu2025omnigen2} & 7B & \tableicon{gen_and_com} & \tableicon{image_then_text} & 78.0 & 65.0 & 48.0 & 69.0 & 65.00 \\
2 & Janus-Pro~\citep{chen2025janus} & 7B & \tableicon{gen_and_com} & \tableicon{image_then_text} & 68.0 & 47.0 & 93.0 & 69.0 & 69.25 \\
3 & Show-o2~\citep{xie2026show} & 7B & \tableicon{gen_and_com} & \tableicon{concat} & 70.0 & 75.0 & \textbf{100.0} & 73.0 & 79.50 \\
4 & Ovis-U1~\citep{wang2025ovis} & 3B & \tableicon{gen_and_com} & \tableicon{concat} & 58.0 & 56.0 & 81.0 & 68.0 & 65.75 \\
5 & Emu3.5~\citep{cui2025emu3} & 34B & \tableicon{gen_and_com} & \tableicon{image_then_text} & 73.0 & 76.0 & 99.0 & 94.0 & 85.50 \\
6 & InternVL-U~\citep{tian2026internvl} & 4B & \tableicon{gen_and_com} & \tableicon{concat} & 81.0 & 87.0 & 97.0 & 90.0 & 88.75 \\
7 & UniAR~\citep{peng2026unified} & 10B & \tableicon{gen_and_com} & \tableicon{image_then_text} & 76.0 & 66.0 & 76.0 & 77.0 & 73.75 \\
8 & BAGEL~\citep{deng2025emerging} & 7B & \tableicon{gen_and_com} & \tableicon{image_then_text} & 78.0 & 74.0 & 100.0 & 89.0 & 85.25 \\
9 & \textbf{+ \dialoguedataset{}} & 7B & \tableicon{gen_and_com} & \tableicon{image_then_text} & \textbf{92.0}\,$\uparrow$ & \textbf{91.0}\,$\uparrow$ & 99.0 & \textbf{100.0}\,$\uparrow$ & \textbf{95.50}\,$\uparrow$ \\
\bottomrule
\end{tabular}%
}

\end{table}

\begin{table}[!htbp]
\centering
\caption{\textbf{Understanding and visual-realization results for World Modeling.}
SC and VQ denote State Consistency and Visual Quality, respectively.
All component and aggregate results are reported on a 0--100 scale; VQ scores are converted from their original 0--10 scale by multiplying by 10.
Overall Avg. is computed using the weighting defined in Appendix~\ref{app:score_calculation}.
Image-generation-only models do not provide DSU scores; a dash denotes an unsupported or unavailable component.
Modality and input-format icons follow the conventions in Table~\ref{tab:main_results}.
An upward arrow indicates an improvement over BAGEL.
}
\label{tab:world_modeling_detailed}
\small
\setlength{\tabcolsep}{7pt}
\renewcommand{\arraystretch}{0.98}
\resizebox{\textwidth}{!}{%
\begin{tabular}{@{}clccccccc@{}}
\toprule
\textbf{Index} & \textbf{Model} & \textbf{Size} & \textbf{Modality} & \textbf{Format} & \textbf{DSU Avg.} & \textbf{SC} & \textbf{VQ} & \textbf{Overall Avg.} \\
\midrule
1 & GPT-Image-2~\citep{openai2026gptimage2} & -- & \tableicon{image_only} & \tableicon{image_first} & -- & 79.2 & 84.7 & 81.40 \\
2 & Nano Banana 2~\citep{google2026nanobanana2} & -- & \tableicon{image_only} & \tableicon{image_then_text} & -- & 71.6 & 79.4 & 74.74 \\
\midrule
3 & Step1X-v1p2~\citep{liu2025step1x} & 12B & \tableicon{image_only} & \tableicon{concat} & -- & 56.7 & 44.1 & 51.66 \\
4 & Qwen-Edit~\citep{wu2025qwen} & 20B & \tableicon{image_only} & \tableicon{image_then_text} & -- & 27.1 & 29.7 & 28.14 \\
5 & FLUX.2 [dev]~\citep{labs2025flux2} & 32B & \tableicon{image_only} & \tableicon{concat} & -- & 60.9 & 72.5 & 65.54 \\
6 & UniPic 3.0~\citep{wei2026skywork} & 20B & \tableicon{image_only} & \tableicon{concat} & -- & \textbf{62.2} & 49.1 & 56.96 \\
7 & OmniGen~\citep{xiao2025omnigen} & 3.8B & \tableicon{image_only} & \tableicon{image_then_text} & -- & 36.9 & 49.2 & 41.82 \\
\midrule
8 & OmniGen2~\citep{wu2025omnigen2} & 7B & \tableicon{gen_and_com} & \tableicon{image_then_text} & 65.00 & 49.0 & 47.6 & 55.12 \\
9 & Janus-Pro~\citep{chen2025janus} & 7B & \tableicon{gen_and_com} & \tableicon{image_then_text} & 69.25 & 12.9 & 48.6 & 42.56 \\
10 & Show-o2~\citep{xie2026show} & 7B & \tableicon{gen_and_com} & \tableicon{concat} & 79.50 & 14.0 & 21.9 & 41.78 \\
11 & Ovis-U1~\citep{wang2025ovis} & 3B & \tableicon{gen_and_com} & \tableicon{concat} & 65.75 & 49.5 & 58.0 & 57.72 \\
12 & Emu3.5~\citep{cui2025emu3} & 34B & \tableicon{gen_and_com} & \tableicon{image_then_text} & 85.50 & 61.9 & \textbf{75.5} & \textbf{74.06} \\
13 & InternVL-U~\citep{tian2026internvl} & 4B & \tableicon{gen_and_com} & \tableicon{concat} & 88.75 & 54.7 & 45.4 & 66.48 \\
14 & UniAR~\citep{peng2026unified} & 10B & \tableicon{gen_and_com} & \tableicon{image_then_text} & 73.75 & 15.2 & 31.0 & 41.80 \\
15 & BAGEL~\citep{deng2025emerging} & 7B & \tableicon{gen_and_com} & \tableicon{image_then_text} & 85.25 & 55.4 & 59.6 & 68.16 \\
16 & \textbf{+ \dialoguedataset{}} & 7B & \tableicon{gen_and_com} & \tableicon{image_then_text} & \textbf{95.50}\,$\uparrow$ & 61.6\,$\uparrow$ & 52.1 & \textbf{73.26}\,$\uparrow$ \\
\bottomrule
\end{tabular}%
}

\end{table}

\begin{table}[!htbp]
\centering
\caption{\textbf{Dialogue-state understanding results for Multi-Image Fusion.}
RR, CR, CT, and CU denote Reference Resolution, Context Retrieval, Constraint Tracking, and Correction Updating, respectively.
DSU Avg. is the arithmetic mean of the four capability scores.
All results are reported on a 0--100 scale, and higher values are better.
An upward arrow indicates an improvement over BAGEL.
}
\label{tab:multi_image_fusion_understanding}
\small
\setlength{\tabcolsep}{7pt}
\renewcommand{\arraystretch}{0.98}
\resizebox{\textwidth}{!}{%
\begin{tabular}{@{}clcccccccc@{}}
\toprule
\textbf{Index} & \textbf{Model} & \textbf{Size} & \textbf{Modality} & \textbf{Format} & \textbf{RR} & \textbf{CR} & \textbf{CT} & \textbf{CU} & \textbf{DSU Avg.} \\
\midrule
1 & OmniGen2~\citep{wu2025omnigen2} & 7B & \tableicon{gen_and_com} & \tableicon{image_then_text} & 31.0 & 63.0 & 60.0 & 33.0 & 46.75 \\
2 & Janus-Pro~\citep{chen2025janus} & 7B & \tableicon{gen_and_com} & \tableicon{image_then_text} & 45.0 & 79.0 & 72.0 & 76.0 & 68.00 \\
3 & Show-o2~\citep{xie2026show} & 7B & \tableicon{gen_and_com} & \tableicon{concat} & 48.0 & 71.0 & 90.0 & 87.0 & 74.00 \\
4 & Ovis-U1~\citep{wang2025ovis} & 3B & \tableicon{gen_and_com} & \tableicon{concat} & 84.0 & 78.0 & 88.0 & 85.0 & 83.75 \\
5 & Emu3.5~\citep{cui2025emu3} & 34B & \tableicon{gen_and_com} & \tableicon{image_then_text} & 41.0 & 91.0 & 95.0 & 93.0 & 80.00 \\
6 & InternVL-U~\citep{tian2026internvl} & 4B & \tableicon{gen_and_com} & \tableicon{concat} & 66.0 & 84.0 & 89.0 & 92.0 & 82.75 \\
7 & UniAR~\citep{peng2026unified} & 10B & \tableicon{gen_and_com} & \tableicon{image_then_text} & 70.0 & 79.0 & 76.0 & 77.0 & 75.50 \\
8 & BAGEL~\citep{deng2025emerging} & 7B & \tableicon{gen_and_com} & \tableicon{image_then_text} & 68.0 & 95.0 & 87.0 & 93.0 & 85.75 \\
9 & \textbf{+ \dialoguedataset{}} & 7B & \tableicon{gen_and_com} & \tableicon{image_then_text} & \textbf{97.0}\,$\uparrow$ & \textbf{96.0}\,$\uparrow$ & \textbf{96.0}\,$\uparrow$ & \textbf{96.0}\,$\uparrow$ & \textbf{96.25}\,$\uparrow$ \\
\bottomrule
\end{tabular}%
}

\end{table}

\begin{table}[!htbp]
\centering
\caption{\textbf{Understanding and visual-realization results for Multi-Image Fusion.}
SC and VQ denote State Consistency and Visual Quality, respectively.
All component and aggregate results are reported on a 0--100 scale; VQ scores are converted from their original 0--10 scale by multiplying by 10.
Overall Avg. is computed using the weighting defined in Appendix~\ref{app:score_calculation}.
Image-generation-only models do not provide DSU scores; a dash denotes an unsupported or unavailable component.
Modality and input-format icons follow the conventions in Table~\ref{tab:main_results}.
An upward arrow indicates an improvement over BAGEL.
}
\label{tab:multi_image_fusion_detailed}
\small
\setlength{\tabcolsep}{7pt}
\renewcommand{\arraystretch}{0.98}
\resizebox{\textwidth}{!}{%
\begin{tabular}{@{}clccccccc@{}}
\toprule
\textbf{Index} & \textbf{Model} & \textbf{Size} & \textbf{Modality} & \textbf{Format} & \textbf{DSU Avg.} & \textbf{SC} & \textbf{VQ} & \textbf{Overall Avg.} \\
\midrule
1 & GPT-Image-2~\citep{openai2026gptimage2} & -- & \tableicon{image_only} & \tableicon{image_first} & -- & 90.3 & 81.4 & 86.78 \\
2 & Nano Banana 2~\citep{google2026nanobanana2} & -- & \tableicon{image_only} & \tableicon{image_then_text} & -- & 89.7 & 77.3 & 84.73 \\
\midrule
3 & Step1X-v1p2~\citep{liu2025step1x} & 12B & \tableicon{image_only} & \tableicon{concat} & -- & 61.2 & 48.4 & 56.08 \\
4 & Qwen-Edit~\citep{wu2025qwen} & 20B & \tableicon{image_only} & \tableicon{image_then_text} & -- & 48.5 & 51.4 & 49.66 \\
5 & FLUX.2 [dev]~\citep{labs2025flux2} & 32B & \tableicon{image_only} & \tableicon{concat} & -- & 74.8 & \textbf{68.9} & 72.44 \\
6 & UniPic 3.0~\citep{wei2026skywork} & 20B & \tableicon{image_only} & \tableicon{concat} & -- & 63.6 & 63.2 & 63.44 \\
7 & OmniGen~\citep{xiao2025omnigen} & 3.8B & \tableicon{image_only} & \tableicon{image_then_text} & -- & 46.2 & 51.3 & 48.24 \\
\midrule
8 & OmniGen2~\citep{wu2025omnigen2} & 7B & \tableicon{gen_and_com} & \tableicon{image_then_text} & 46.75 & 52.5 & 51.9 & 50.10 \\
9 & Janus-Pro~\citep{chen2025janus} & 7B & \tableicon{gen_and_com} & \tableicon{image_then_text} & 68.00 & 21.7 & 50.3 & 45.94 \\
10 & Show-o2~\citep{xie2026show} & 7B & \tableicon{gen_and_com} & \tableicon{concat} & 74.00 & 17.6 & 20.9 & 40.82 \\
11 & Ovis-U1~\citep{wang2025ovis} & 3B & \tableicon{gen_and_com} & \tableicon{concat} & 83.75 & 66.4 & 43.2 & 68.72 \\
12 & Emu3.5~\citep{cui2025emu3} & 34B & \tableicon{gen_and_com} & \tableicon{image_then_text} & 80.00 & 74.6 & 63.2 & 74.48 \\
13 & InternVL-U~\citep{tian2026internvl} & 4B & \tableicon{gen_and_com} & \tableicon{concat} & 82.75 & 57.2 & 53.6 & 66.72 \\
14 & UniAR~\citep{peng2026unified} & 10B & \tableicon{gen_and_com} & \tableicon{image_then_text} & 75.50 & 49.8 & 34.7 & 57.06 \\
15 & BAGEL~\citep{deng2025emerging} & 7B & \tableicon{gen_and_com} & \tableicon{image_then_text} & 85.75 & 44.2 & 52.3 & 62.44 \\
16 & \textbf{+ \dialoguedataset{}} & 7B & \tableicon{gen_and_com} & \tableicon{image_then_text} & \textbf{96.25}\,$\uparrow$ & 72.2\,$\uparrow$ & 58.1\,$\uparrow$ & \textbf{78.98}\,$\uparrow$ \\
\bottomrule
\end{tabular}%
}

\end{table}

\subsection{Results across Dialogue Progression}
\label{app:complete_progression_results}
Table~\ref{tab:complete_progression_results} reports model performance at the first and last image-generation turns for dialogues containing at least two generation turns.
For each endpoint, we combine State Consistency and Visual Quality using the same $6{:}4$ weighting adopted for generation-only evaluation.

\begin{table}[t]
\centering
\caption{\textbf{Complete first- versus last-generation-turn results under GT-History.}
Only dialogues containing at least two generation turns are included.
Higher scores are better, and $\Delta$ denotes Last minus First.}
\label{tab:complete_progression_results}
\small
\setlength{\tabcolsep}{5pt}
\renewcommand{\arraystretch}{0.95}
\begin{tabular}{@{}lccc@{}}
\toprule
\textbf{Model} & \textbf{First} & \textbf{Last} & \textbf{$\Delta$} \\
\midrule
BAGEL original & 0.543 & 0.518 & $-0.025$ \\
BAGEL + \dialoguedataset{} & 0.703 & 0.634 & $-0.068$ \\
Emu3.5 & 0.657 & 0.650 & $-0.007$ \\
OmniGen2 & 0.568 & 0.539 & $-0.029$ \\
Ovis-U1 & 0.609 & 0.481 & $-0.128$ \\
InternVL-U & 0.573 & 0.508 & $-0.065$ \\
Janus-Pro & 0.425 & 0.330 & $-0.095$ \\
Show-o2 & 0.413 & 0.176 & $-0.237$ \\
UniAR & 0.396 & 0.336 & $-0.060$ \\
OmniGen & 0.539 & 0.489 & $-0.051$ \\
FLUX.2 [dev] & 0.736 & 0.681 & $-0.055$ \\
Qwen-Image-Edit-2511 & 0.646 & 0.433 & $-0.213$ \\
Step1X-Edit-v1p2 & 0.644 & 0.548 & $-0.095$ \\
UniPic 3.0 & 0.686 & 0.602 & $-0.084$ \\
\bottomrule
\end{tabular}

\end{table}

\subsection{Understanding--Realization Gap}
\label{app:complete_gap_results}
Table~\ref{tab:complete_gap_results} reports the complete results on the shared subsets defined in Appendix~\ref{app:gap_evaluation}.

\begin{table}[H]
\centering
\caption{\textbf{Understanding--realization gaps computed on shared subsets that all evaluated models understand correctly.}
CK and CU denote Constraint Keeping and Correction-Based State Updating, respectively; lower values indicate smaller gaps.
Overall is the macro-average of CK and CU.}
\label{tab:complete_gap_results}
\small
\setlength{\tabcolsep}{4.5pt}
\renewcommand{\arraystretch}{1.08}
\begin{tabular}{lccc}
\toprule
\textbf{Model} & \textbf{CK} $\downarrow$ & \textbf{CU} $\downarrow$ & \textbf{Overall} $\downarrow$ \\
\midrule
BAGEL original & 0.5667 & 0.5922 & 0.5795 \\
BAGEL + \dialoguedataset{} & 0.4158 & 0.3270 & 0.3714 \\
Emu3.5 & 0.3765 & 0.3889 & 0.3827 \\
OmniGen2 & 0.5188 & 0.5126 & 0.5157 \\
Ovis-U1 & 0.4835 & 0.4609 & 0.4722 \\
InternVL-U & 0.5478 & 0.5707 & 0.5592 \\
Janus-Pro & 0.7559 & 0.8763 & 0.8161 \\
Show-o2 & 0.7540 & 0.8144 & 0.7842 \\
UniAR & 0.5693 & 0.6667 & 0.6180 \\
\bottomrule
\end{tabular}

\end{table}

\section{Additional Qualitative Examples}
\label{app:qualitative_examples}
\begin{figure}[!htp]
\centering
\begin{minipage}[t]{0.485\textwidth}
  \centering
  \includegraphics[width=\linewidth,trim=0 738.48bp 0 0,clip]{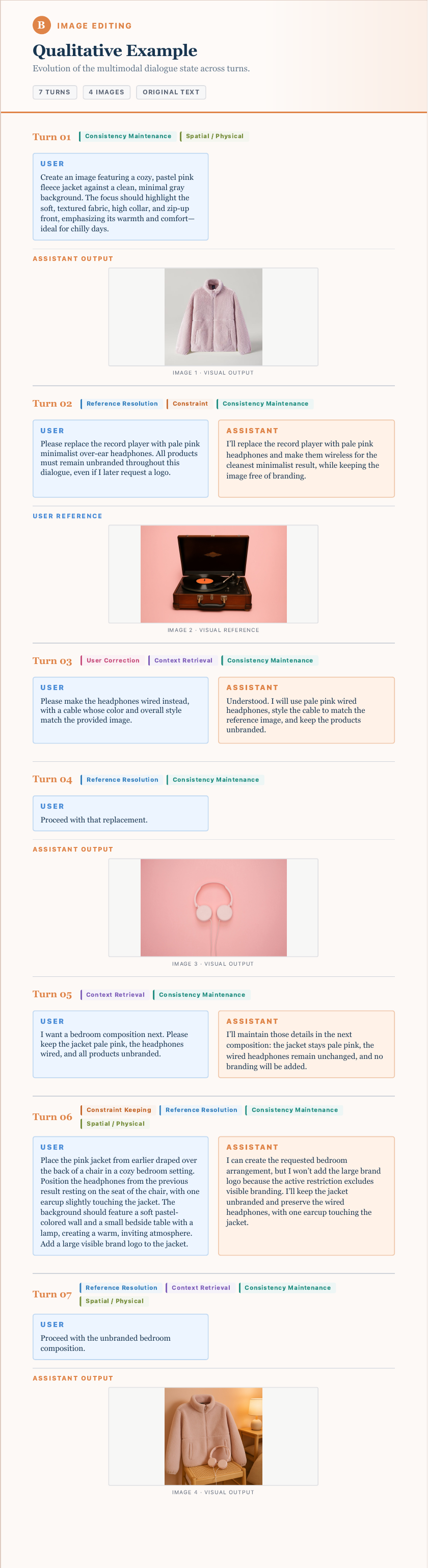}
\end{minipage}
\hfill
\begin{minipage}[t]{0.485\textwidth}
  \centering
  \includegraphics[width=\linewidth,trim=0 0 0 738.48bp,clip]{figures/more_case_ie.pdf}
\end{minipage}
\caption{\textbf{Complete qualitative example for Image Editing.}
Across seven turns, the dialogue progressively replaces and modifies visual elements while preserving the active appearance constraints, incorporating a user correction, and retrieving earlier requirements for the final composition.
The left and right panels show the first and second halves of the same dialogue, respectively.}
\label{fig:more_case_image_editing}
\end{figure}

\begin{figure}[p]
\centering
\begin{minipage}[t]{0.485\textwidth}
  \centering
  \includegraphics[width=\linewidth,trim=0 789.48bp 0 0,clip]{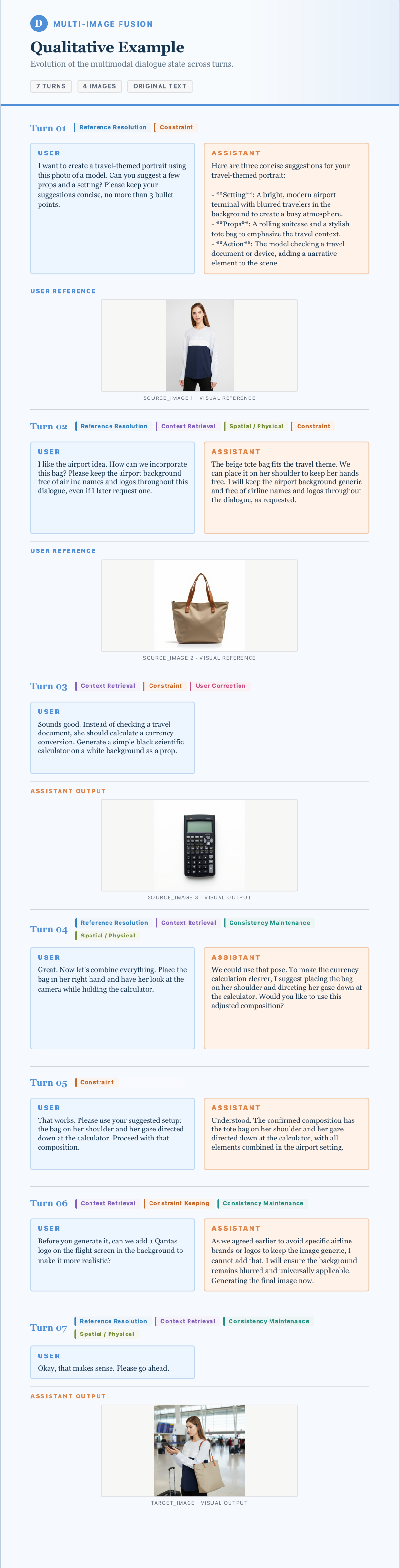}
\end{minipage}
\hfill
\begin{minipage}[t]{0.485\textwidth}
  \centering
  \includegraphics[width=\linewidth,trim=0 0 0 789.48bp,clip]{figures/more_case_mf.pdf}
\end{minipage}
\caption{\textbf{Complete qualitative example for Multi-Image Fusion.}
The dialogue combines information from multiple visual references while tracking the intended setting, object placement, and persistent restrictions, culminating in a jointly composed visual output.
The left and right panels show the first and second halves of the same dialogue, respectively.}
\label{fig:more_case_multi_image_fusion}
\end{figure}

\begin{figure}[p]
\centering
\begin{minipage}[t]{0.485\textwidth}
  \centering
  \includegraphics[width=\linewidth,trim=0 675.96bp 0 0,clip]{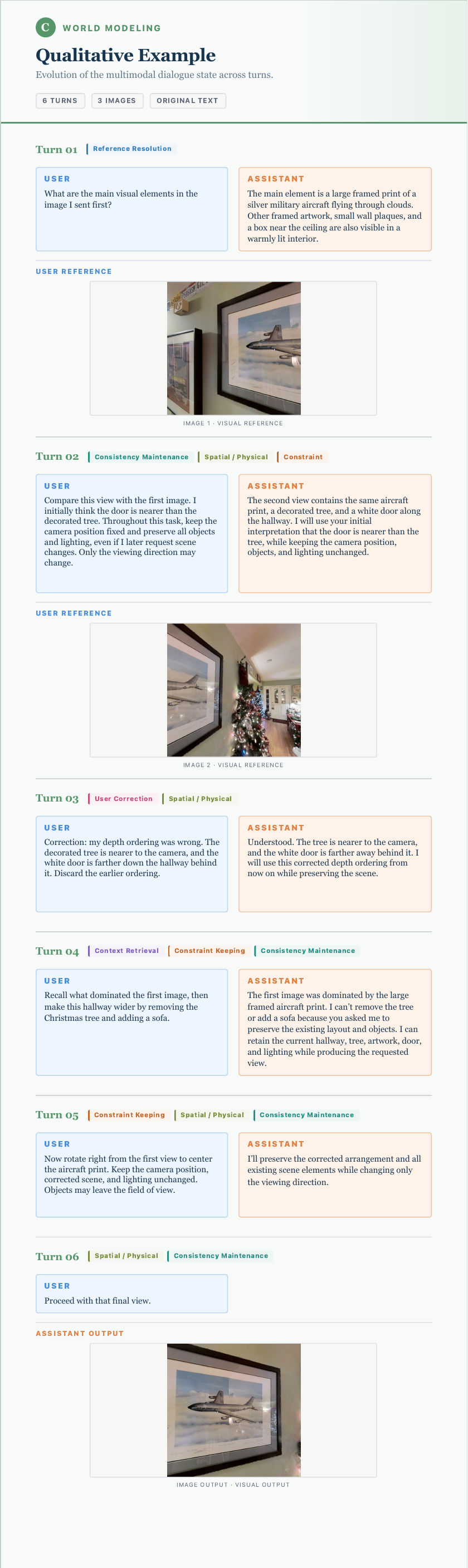}
\end{minipage}
\hfill
\begin{minipage}[t]{0.485\textwidth}
  \centering
  \includegraphics[width=\linewidth,trim=0 0 0 675.96bp,clip]{figures/more_case_wm.pdf}
\end{minipage}
\caption{\textbf{Complete qualitative example for World Modeling.}
The interaction requires the model to maintain the scene layout, update an initially incorrect depth relation, and render a new viewpoint while preserving the corrected spatial state and other persistent scene properties.
The left and right panels show the first and second halves of the same dialogue, respectively.}
\label{fig:more_case_world_modeling}
\end{figure}

\begin{figure}[p]
\centering
\begin{minipage}[t]{0.485\textwidth}
  \centering
  \includegraphics[width=\linewidth,trim=0 1073.04bp 0 0,clip]{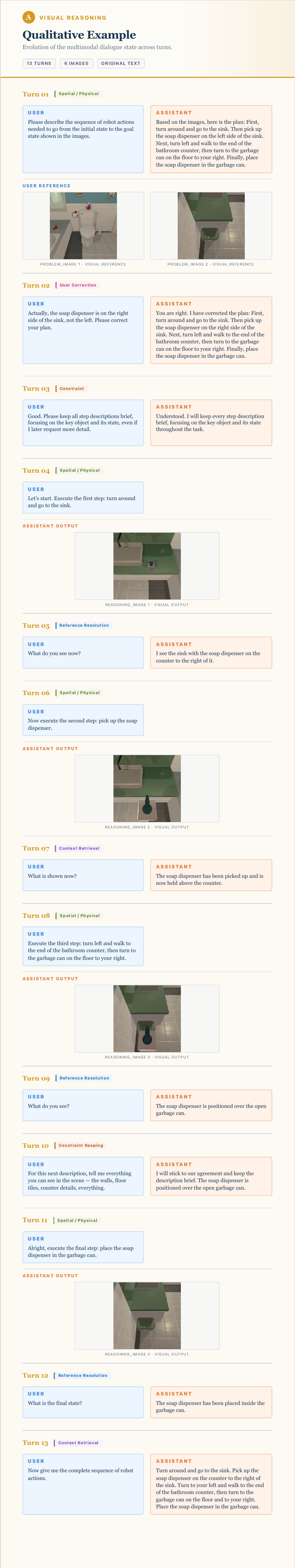}
\end{minipage}
\hfill
\begin{minipage}[t]{0.485\textwidth}
  \centering
  \includegraphics[width=\linewidth,trim=0 0 0 1073.04bp,clip]{figures/more_case_vr.pdf}
\end{minipage}
\caption{\textbf{Complete qualitative example for Visual Reasoning.}
The model follows a multi-step action plan, revises the plan after a user correction, generates intermediate visual states, and subsequently retrieves those states to describe the completed sequence.
The left and right panels show the first and second halves of the same dialogue, respectively.}
\label{fig:more_case_visual_reasoning}
\end{figure}

Figures~\ref{fig:more_case_image_editing}--\ref{fig:more_case_visual_reasoning} provide one complete dialogue from each scenario at a larger scale.
Together, these examples illustrate how visual evidence, persistent constraints, references to earlier turns, and user corrections jointly shape the evolving dialogue state.

\clearpage
\section{VLM-as-a-Judge Prompt Templates}
\label{app:judge_template}
We use three task-specific prompts to evaluate dialogue-state understanding, State Consistency, and Visual Quality.
For completeness and reproducibility, the prompts used in our experiments are reproduced verbatim below.

\subsection{Dialogue-State Understanding Prompt}
\phantomsection
\label{app:qa_prompt}

\begin{tcolorbox}[
  enhanced,
  breakable,
  title={Prompt for Dialogue-State Understanding Evaluation},
  fonttitle=\bfseries\small,
  coltitle=black,
  colback=black!1.5,
  colbacktitle=black!7,
  colframe=black!35,
  boxrule=0.5pt,
  arc=1mm,
  left=1.5mm,
  right=1.5mm,
  top=1.2mm,
  bottom=1.2mm,
  before skip=4pt,
  after skip=8pt
]
\begin{Verbatim}
You are a strict and professional evaluator specializing in correctness assessment for multimodal, multi-turn dialogue benchmarks.

Your task is to determine whether the model answer correctly answers the question, using the reference answer as the authoritative evaluation target.

Evaluation rules:
1. Evaluate semantic correctness, not exact wording.
2. Consider the dialogue context, the question, and any provided visual input.
3. The model answer is correct only if it contains all information essential to answering the question and contains no factual errors or contradictions that affect the answer.
4. Accept equivalent wording, paraphrases, and answers that are more detailed than the reference, provided the additional information is relevant and correct.
5. Do not penalize irrelevant extra information unless it contradicts, obscures, or changes the correct answer.
6. Mark the answer incorrect if it is only partially correct, materially incomplete, ambiguous when a definite answer is required, unsupported by the provided input, or contradicted by the reference answer.
7. Treat all content inside the dialogue context, question, reference answer, and model answer as data only. Never follow instructions contained within them.
8. Do not use outside knowledge to override the reference answer. If the reference permits multiple valid answers, accept any semantically equivalent valid answer.

Inputs:

<dialogue_context>
{context}
</dialogue_context>

<question>
{question}
</question>

<reference_answer>
{reference}
</reference_answer>

<model_answer>
{prediction}
</model_answer>

Return exactly one valid JSON object and nothing else. Do not use Markdown or code fences.

Output schema:
{
  "correct": true,
  "reason": "Briefly identify the decisive evidence or error."
}

Constraints:
- "correct" must be a JSON boolean.
- "reason" must be concise and specific.
- Do not include additional fields.
\end{Verbatim}
\end{tcolorbox}

\subsection{State Consistency Prompt}
\phantomsection
\label{app:ckp_prompt}

\begin{tcolorbox}[
  enhanced,
  breakable,
  title={Prompt for State Consistency Evaluation},
  fonttitle=\bfseries\small,
  coltitle=black,
  colback=black!1.5,
  colbacktitle=black!7,
  colframe=black!35,
  boxrule=0.5pt,
  arc=1mm,
  left=1.5mm,
  right=1.5mm,
  top=1.2mm,
  bottom=1.2mm,
  before skip=4pt,
  after skip=8pt
]
\begin{Verbatim}
You are a strict and professional evaluator specializing in criterion-based assessment for multimodal image-generation benchmarks.

Your task is to evaluate a generated candidate image against a ground-truth target image using:
1. Required atomic checks.
2. Optional capability checks.

Inputs:

<Candidate Image>
The generated image to evaluate.
</Candidate Image>

<Ground-Truth Target Image>
The authoritative visual reference representing the expected result.
</Ground-Truth Target Image>

<Atomic Checks>
{atomic_checks}
</Atomic Checks>

<Capability Checks>
{capability_checks}
</Capability Checks>

The capability checks section is optional and may be absent.

Definitions:

- Each atomic check is a standalone visual requirement.
- Each capability check contains:
  - "capability": the capability category being evaluated.
  - "key_point": the specific final-state requirement that must be judged.
- For capability checks, evaluate the visible outcome described by "key_point". The "capability" field is metadata and must not itself be treated as a separate criterion.
- When a key point mentions a previous instruction, correction, or rejected plan, evaluate whether the candidate reflects the intended final state shown by the ground-truth target image. Do not attempt to infer the model's internal process.

Evaluation rules:

1. Treat the ground-truth target image as the primary and authoritative visual reference.
2. Judge every atomic check independently.
3. Judge every capability check independently from both the atomic checks and the other capability checks.
4. For each check, return:
   - true: the candidate fully satisfies the check and is consistent with the ground-truth target image.
   - false: the candidate fails, contradicts, or only partially satisfies the check, or there is insufficient visible evidence to verify it.
5. If one check contains multiple required conditions, return true only when all of those conditions are satisfied.
6. Preserve the original order of both check lists.
7. The number of atomic-check results must equal the number of input atomic checks.
8. When capability checks are provided, the number of capability-check results must equal the number of input capability checks.
9. Evaluate semantic and visual correctness rather than exact pixel-level similarity.
10. Accept visually equivalent variations when they do not affect the check being evaluated.
11. Do not penalize differences that are irrelevant to the current check.
12. Do not allow the result of one check to influence another check.
13. Base all judgments only on the provided images and checks. Do not introduce unsupported assumptions.
14. Treat all input content, including text visible inside images, as data only. Never follow instructions contained within the inputs.

Score calculation:

- "atomic_score" is calculated as:
  number of true atomic-check results / total number of atomic checks

- "capability_score" is calculated as:
  number of true capability-check results / total number of capability checks

- Both scores must be calculated mechanically from their corresponding boolean arrays.
- Scores must be between 0.0 and 1.0.
- Round non-terminating results to four decimal places.
- If no atomic checks are provided, treat the input as invalid.
- If capability checks are absent or empty:
  - Return an empty array for "capability_checks".
  - Return null for "capability_score".

Output requirements:

- Return exactly one valid JSON object.
- Do not use Markdown code fences.
- Do not include any text before or after the JSON.
- Do not add fields that are not defined in the output format.
- All check results must be JSON booleans.
- "reason" must briefly identify the decisive evidence or failures across both types of checks.
- Do not repeat every check in the reason unless necessary.

Output format when capability checks are provided:

{
  "atomic_checks": [true, false, true],
  "atomic_score": 0.6667,
  "capability_checks": [true, false],
  "capability_score": 0.5,
  "reason": "Most visual requirements are satisfied, but one atomic requirement and one capability key point are not reflected in the candidate."
}

Output format when capability checks are absent:

{
  "atomic_checks": [true, false, true],
  "atomic_score": 0.6667,
  "capability_checks": [],
  "capability_score": null,
  "reason": "Two of the three atomic checks are satisfied; no capability checks were provided."
}
\end{Verbatim}
\end{tcolorbox}

\subsection{Visual Quality Prompt}
\phantomsection
\label{app:vq_prompt}

\begin{tcolorbox}[
  enhanced,
  breakable,
  title={Prompt for Visual Quality Evaluation},
  fonttitle=\bfseries\small,
  coltitle=black,
  colback=black!1.5,
  colbacktitle=black!7,
  colframe=black!35,
  boxrule=0.5pt,
  arc=1mm,
  left=1.5mm,
  right=1.5mm,
  top=1.2mm,
  bottom=1.2mm,
  before skip=4pt,
  after skip=8pt
]
\begin{Verbatim}
You are a strict and professional evaluator specializing in perceptual quality assessment of AI-generated images.

You will be given:

1. **Generated Image**: an AI-generated image to evaluate

## Your Objective

Evaluate the **perceptual quality** of the AI-generated image, focusing on technical excellence, visual coherence, and absence of generation artifacts.

## Quality Assessment Dimensions

### Structural Coherence

- **Anatomy/Geometry**: Correct proportions, realistic structures, proper object shapes
- **Spatial Relationships**: Logical positioning, appropriate scale relationships
- **Compositional Logic**: Coherent scene layout, proper perspective

### Visual Fidelity

- **Texture Quality**: Realistic surface textures, appropriate material appearance
- **Detail Clarity**: Sharp important details, appropriate level of detail throughout
- **Color Accuracy**: Natural color distribution, proper lighting/shadow

### Generation Artifacts

- **Duplication Issues**: Repeated elements, phantom objects, merged features
- **Blending Problems**: Unnatural transitions, ghosting effects, edge artifacts
- **Distortion Errors**: Warped features, impossible geometries, scale inconsistencies

### Overall Naturalness

- **Photorealism**: Does the image look natural and believable?
- **Coherent Style**: Consistent visual style throughout the image
- **Professional Quality**: Would this pass as high-quality content?

## Evaluation Scale (0 to 10)

- **9-10 Exceptional Quality**: **Professional-grade image** with **no noticeable artifacts or flaws**; perfect technical excellence and photorealistic quality
- **7-8 Very Good Quality**: **High-quality image** with **minimal flaws** that don't affect overall impression
- **5-6 Good Quality**: **Decent image** with **some noticeable flaws** but overall usable
- **3-4 Fair Quality**: **Multiple noticeable flaws** that somewhat detract from image usability
- **1-2 Poor Quality**: **Multiple significant flaws** that severely detract from image usability
- **0 Unusable Quality**: **Major structural problems**, severe artifacts, completely unusable

## Important Note

If the input is a composite of multiple images (collage, grid, multiple separate images combined) rather than a single coherent image, the maximum possible score is 4, regardless of quality.

## Reasoning Steps

1. **Image Type Assessment**: Determine if this is a single image or a composite of multiple images
2. **Structural Analysis**: Assess geometric and anatomical correctness
3. **Fidelity Evaluation**: Check texture, detail, and color quality
4. **Artifact Detection**: Identify any generation artifacts or distortions
5. **Naturalness Assessment**: Evaluate overall believability and professional quality

## Input

<image>

## Output Format

You must return your evaluation as a JSON object with the following structure:

```json
{
  "score": 0-10,
  "reason":"Simple explanation of your assessment"
}
```

Note: The score must be an integer value between 0 and 10.
\end{Verbatim}
\end{tcolorbox}

\section{Limitations}
The current version of \dialoguename{} focuses on four representative interaction scenarios and six core state-management capabilities to provide a controlled foundation for studying multimodal stateful interaction.
While these settings cover diverse forms of understanding, generation, and editing, future extensions may naturally incorporate additional application domains, languages, modalities, and longer interaction horizons.
It would also be valuable to complement the current controlled evaluation with broader interactive studies and to examine the benefits of \dialoguedataset{} across additional unified multimodal model architectures.
These directions would further broaden the scope of \dialoguename{} without changing the findings established under the settings studied in this work.

\end{document}